\documentclass[times, review, 10pt]{elsarticle}
\usepackage{adjustbox}

\usepackage{amssymb}
\usepackage{amsmath}
\usepackage{makecell}
\usepackage{graphicx}

\journal{Pattern Recognition}

\begin{document}

\begin{frontmatter}



\title{Fine-Grained Visual Preprocessing and Dual-Stream Temporal Modeling for Multimodal Sentiment Analysis on Social Media}


\author[1]{Su Li}
\ead{2579864351@qq.com}
\author[1]{Yigong Zhang\corref{cor1}}
\ead{zhyg@kmu.edu.cn}

\author[1]{Lei Xiong}
\ead{xlei0320@163.com}
\author[1]{Chune Li}
\ead{lichune88@163.com}
\cortext[cor1]{Corresponding Author}

\affiliation[1]{
    organization={School of Information Engineering, Kunming University},
    addressline={No. 2 Puxin Road, Guandu District, Yangpu Campus},
    city={Kunming},
    postcode={650214},
    state={Yunnan},
    country={China}
}

\begin{abstract}
Multimodal sentiment analysis often remains text-dominant due to raw-video noise and insufficient temporal modeling. Using CH-SIMS v2.0S, this study proposes three improvements: the NAPS pipeline---a seven-stage system integrating face tracking,identity embedding, and normalized lip-motion analysis to reduce visual noise;DS-TANet, combining an EfficientNetB2 static stream, RAFT optical-flow motion stream, motion-guided attention, and Bi-GRU temporal modeling; and DS-TAFNet, fusing visual and MacBERT-Base textual representations via concatenation fusion. With NAPS, the static visual baseline achieves 80.98\% Macro F1, comparable to the text baseline of 80.55\%; DS-TANet improves visual Macro F1 to 82.58\%;and DS-TAFNet achieves 87.49\% accuracy and 87.48\% Macro F1. These results demonstrate that improving visual input quality and temporal representation is more
effective than increasing fusion complexity under limited-data conditions.
\end{abstract}

\begin{keyword}
Multimodal Sentiment Analysis \sep Visual Preprocessing \sep Dual-Stream Network \sep Optical Flow \sep Multimodal Fusion

\end{keyword}

\end{frontmatter}



\section{Introduction}

\label{sec1}

Sentiment analysis identifies the emotional polarity expressed in text, speech, or video, and underpins applications ranging from social-media monitoring to human computer interaction. With the rapid growth of short-video platforms, text-only methods have become insufficient for real-world emotion understanding: in natural communication, affect is often conveyed through tone, facial movement, and gesture, or through their inconsistency with verbal content. Multimodal Sentiment Analysis (MSA) therefore integrates textual, visual, and acoustic cues to obtain a more complete representation of affective information~\cite{das2023survey}.

Existing MSA methods nevertheless remain text-dominant~\cite{wang2023tetfn}, with the visual modality contributing comparatively little and thereby limiting the effectiveness of multimodal fusion~\cite{peng2022ogm}. Two causes stand out. At the data level, most studies feed raw video clips directly into the model, so visual features are contaminated by multi-person co-occurrence, non-target-speaker interference, profile faces, occlusion, and unstable illumination. At the modeling level, static frame-level features still dominate, leaving temporal facial dynamics largely unexploited, although the continuous changes of eyebrows, eyes, and lips carry important affective cues.

To address these issues, this paper proposes a multimodal sentiment analysis framework for social media video scenarios and evaluates it on CH-SIMS v2.0S ~\cite{liu2022chsimsv2}. The dataset contains natural conversational clips from films, television series, and variety shows, covering visual conditions such as multi-person co-occurrence, shot changes, pose variation, complex illumination, and spontaneous facial expressions. These characteristics are similar to the complex natural conversational scenes commonly found in social media videos, making the dataset suitable for evaluating target-speaker localization, visual noise reduction, and temporal facial modeling. The main contributions of this paper are summarized as follows:

(1) NAPS, a visual preprocessing pipeline that removes raw-video noise at the source, combining normalized lip-motion analysis with face tracking and identity representation to localize the target speaker and improve the reliability of visual inputs.

(2) DS-TANet, a dual-stream temporal visual encoder that jointly models static facial appearance and dynamic motion through EfficientNetB2, RAFT optical flow, motion-guided attention, and GRU-based temporal modeling.

(3) DS-TAFNet, a multimodal framework integrating DS-TANet with a prompt-tuned MacBERT text encoder, together with a comparative study of fusion strategies that clarifies the role of lightweight fusion under small-dataset conditions.

The remainder of this paper is organized as follows: Section 2 reviews related work; Section 3 presents the dataset and the NAPS pipeline; Section 4 describes DS-TANet and DS-TAFNet; Section 5 presents the experimental results and analysis;Section 6 concludes the paper.

\section{Related Work}
\label{sec2}

\subsection{Multimodal Sentiment Analysis}
\label{subsec2.1}
Sentiment analysis has evolved from text-based unimodal approaches to multimodal joint modeling, supported by benchmarks such as CMU-MOSEI~\cite{zadeh2018mosei}. For Chinese, CH-SIMS provides 2,281 video clips with separate textual, visual, acoustic, and multimodal annotations, allowing the affective information carried by each modality to be evaluated independently~\cite{yu2020ch}. Liu et al. extended it to CH-SIMS v2.0, enlarging the supervised subset to 4,402 clips and deliberately increasing linguistic ambiguity so that sentiment cannot be inferred from text alone, which encourages the exploitation of visual and acoustic cues~\cite{liu2022chsimsv2}. Notably, on weak-sentiment samples the visual modality reaches an Acc2\_weak of 68.79\%, exceeding the 62.49\% of the textual modality, indicating that non-verbal information provides complementary evidence and can be more discriminative when linguistic cues are weak or ambiguous. Consistently, Das and Singh identify strengthening the effective contribution of the visual modality as a central challenge in MSA~\cite{das2023survey}.

\subsection{Visual Affective Feature Extraction}
\label{subsec2.2}
Visual affective feature extraction has shifted from costly and poorly generalizing handcrafted features, such as facial action units and landmark displacements, to deep representations. Pretrained CNNs became standard for single-frame static features, and EfficientNet~\cite{tan2019efficientnet} reached state-of-the-art ImageNet accuracy with far fewer parameters via compound scaling, becoming a mainstream efficient backbone. Frame-level features, however, cannot capture the temporal dynamics of emotion: continuous movements such as eyebrow fluctuations and gradual lip-corner changes carry richer affective signals than any single frame, yet remain underexplored in MSA~\cite{liu2023expression}. Recent facial expression recognition work addresses this limitation by adaptively fusing static appearance features with motion-correlation cues~\cite{li2023multiscale}. More recent work further improves robustness under low-resolution and occluded conditions through multiscale contextual fusion~\cite{gan2025context}.Optical flow offers an effective pixel-level dynamic cue, and RAFT~\cite{teed2020raft} substantially advanced flow-estimation accuracy, but its use for temporal facial affective modeling is still limited. Together with the two-stream network of Simonyan and Zisserman~\cite{simonyan2014twostream}, which pioneered parallel RGB-spatial and optical-flow-temporal modeling, and recent Pattern Recognition work that revisits the contribution of temporal versus spatial features through parallel cross-fusion~\cite{liu2026temporal},these lines of work motivate the dual-stream temporal design adopted in DS-TANet.

\subsection{Multimodal Fusion Strategies}
\label{subsec2.3}
Multimodal fusion falls into three paradigms. Early fusion concatenates modality features before classification, which is simple but sensitive to modality-quality imbalance; late fusion combines per-modality predictions, preserving modality independence while discarding cross-modal interactions; and intermediate fusion, now dominated by Transformer-based cross-modal attention~\cite{tsai2019multimodal}, achieves fine-grained alignment and strong gains on large-scale datasets. Fusion effectiveness, however, depends not only on the fusion module but also on unimodal feature quality and training-data scale: when features are weak or samples too few for a complex module to converge, fusion may yield no stable gain or even add noise~\cite{wei2026visual}. This compatibility between data scale and fusion complexity has rarely been analyzed systematically on small-scale Chinese datasets.

\subsection{Chinese Pretrained Language Models}
\label{subsec2.4}
For the text modality, BERT-series models dominate NLP~\cite{devlin2019bert}. MacBERT~\cite{cui2020macbert}, built on RoBERTa, replaces BERT's masking with a correction-oriented objective based on synonym substitution, reducing the pretrain--finetune input mismatch and achieving state-of-the-art results on several Chinese NLP benchmarks, which makes it a robust backbone for Chinese MSA.

\subsection{Research Gaps}
\label{subsec2.5}
Three limitations therefore remain: visual preprocessing is often neglected, so raw-video noise contaminates feature extraction; visual modeling still relies on static single-frame features, underusing facial temporal dynamics; and fusion strategies are rarely analyzed on small-scale Chinese datasets. Sections 3 and 4 address all three.

\section{Dataset and NAPS Preprocessing Pipeline}
\label{sec3}
\subsection{CH-SIMS v2.0S Dataset}
\label{subsec3.1}

This study uses CH-SIMS v2.0S, the supervised subset of CH-SIMS v2.0, obtained from the official \textit{thuiar/ch-sims-v2} release. Although the original paper reports 4,402 supervised samples, the official training, validation, and test splits contain 2,722, 647, and 1,034 samples, totaling 4,403; the statistics reported here therefore follow the current official release.

Samples are labeled positive, neutral, or negative. Neutral samples have ambiguous semantic boundaries in polarity classification and account for only 12.1\% of the data, so retaining them as a separate class would aggravate class imbalance and increase ambiguity at the class boundaries. This study therefore formulates the task as positive--negative binary classification and excludes the 533 neutral samples; four clips in which the speaker remained in a profile-face orientation throughout were also removed, as no reliable visual features could be extracted. This leaves 3,866 valid samples, divided according to the official \textit{mode} field as summarized in Table~\ref{t1}. Because negative samples slightly outnumber positive ones, balanced sampling and Focal Loss were adopted during training.

\begin{table}[h]
\centering
\begin{tabular}{l c c c}
\hline
Subset & Positive & Negative & Total \\
\hline
Training   & 1,092 & 1,308 & 2,400 \\
Validation &   230 &   325 &   555 \\
Test       &   423 &   488 &   911 \\
\hline
Total      & 1,745 & 2,121 & 3,866 \\
\hline
\end{tabular}
\caption{Sample distribution of CH-SIMS v2.0S after preprocessing}\label{t1}
\end{table}

\subsection{NAPS Preprocessing Pipeline}
\label{subsec3.2}

Raw MSA videos often contain visual interference, such as multiple persons, non-target speakers, profile faces, and occlusions. To improve input quality, this study develops NAPS (Normalized Adaptive Pose-robust Speaker Localization), a seven-stage automated pipeline with a low-light recovery mode. NAPS localizes the target speaker and converts raw conversational videos into $224\times224$ speaker-centered facial clips for downstream affective modeling.

\begin{figure}[t]
\centering
\makebox[\linewidth][c]{%
    \includegraphics[
        width=1.05\linewidth,
        keepaspectratio
    ]{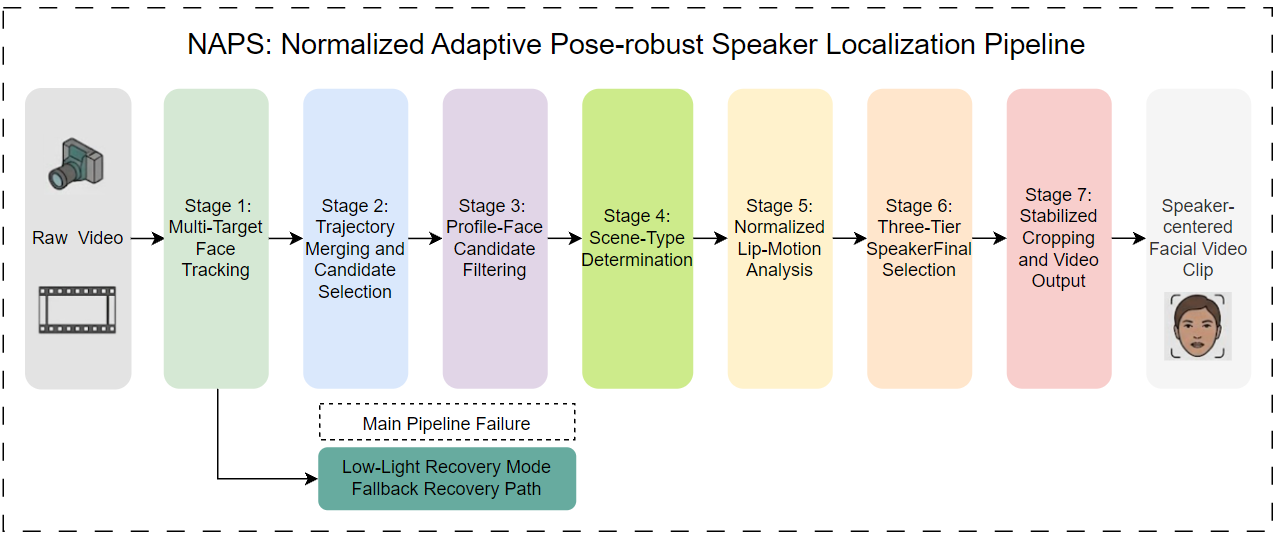}%
}
\caption{NAPS: Normalized Adaptive Pose-robust Speaker Localization Pipeline.}
\label{f1}
\end{figure}

The overall pipeline is shown in Fig.~\ref{f1}. Its main thresholds were selected through preliminary experiments on the training and validation sets to balance detection recall, false-positive suppression, and trajectory stability.

\par\textbf{Stage 1: Multi-Target Face Tracking.}
MediaPipe~\cite{lugaresi2019mediapipe} detects faces in each frame with a confidence threshold of $\geq 0.55$ and outputs bounding boxes, nose tips, and mouth centers. YCrCb skin-color filtering removes false-positive non-face regions, while an InceptionResNetV1 model pretrained on VGGFace2~\cite{cao2018vggface2} extracts 512-dimensional L2-normalized face embeddings for each detected face region.

\begin{figure}[t]
\centering
\makebox[\linewidth][c]{%
    \includegraphics[
        width=1\linewidth,
        keepaspectratio
    ]{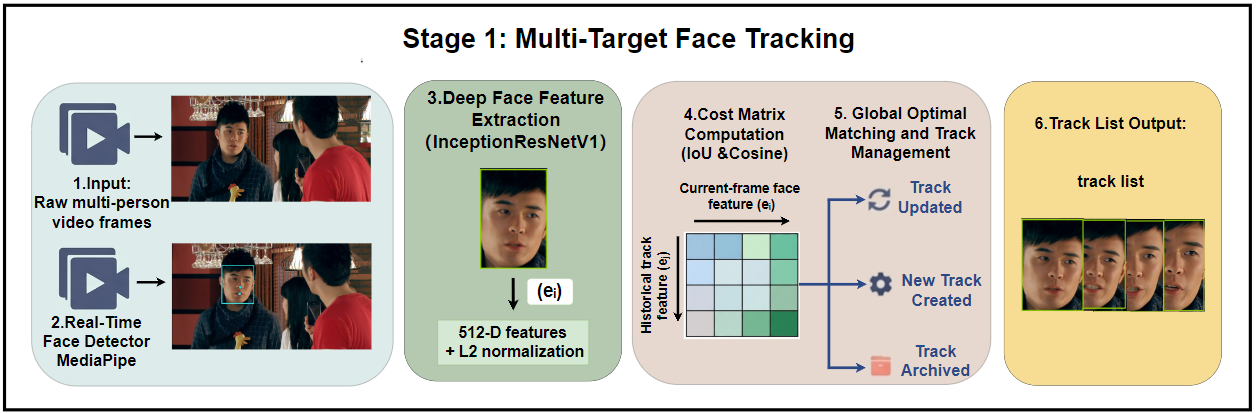}%
}
\caption{Technical pipeline of Stage 1: Multi-Target Face Tracking.}
\label{f2}
\end{figure}

For cross-frame identity association, a cost matrix is constructed between consecutive frames, and the Hungarian algorithm is used to obtain the globally optimal bipartite matching. The matching cost combines spatial overlap and appearance similarity:

\begin{equation}
\text{cost}(i,j) = 1 - \bigl[
0.4\times\operatorname{IoU}(b_i,b_j)
+ 0.6\times\operatorname{cosine}(e_i,e_j)
\bigr]
\label{eq:match_cost}
\end{equation}

\noindent where $b_i$ and $b_j$ are face bounding boxes in consecutive frames, and $e_i$ and $e_j$ are their corresponding embeddings. IoU measures spatial overlap, whereas cosine similarity measures appearance-based identity consistency. This combination of spatial association and appearance-based identity cues is consistent with recent visual multi-object tracking methods that employ re-identification to maintain identity continuity under occlusion and temporary disappearance~\cite{vanma2024mot}.A match is accepted when $(1-\text{cost})>0.50$. If $\operatorname{IoU}>0.5$, the pairwise cost is set to zero, giving the match the highest priority in the Hungarian assignment. Tracks missing for more than 15 consecutive frames are terminated and archived. The pipeline is shown in Fig.~\ref{f2}.

\par
\begin{figure}[t]
\centering
\makebox[\linewidth][c]{%
    \includegraphics[
        width=0.9\linewidth,
        keepaspectratio
    ]{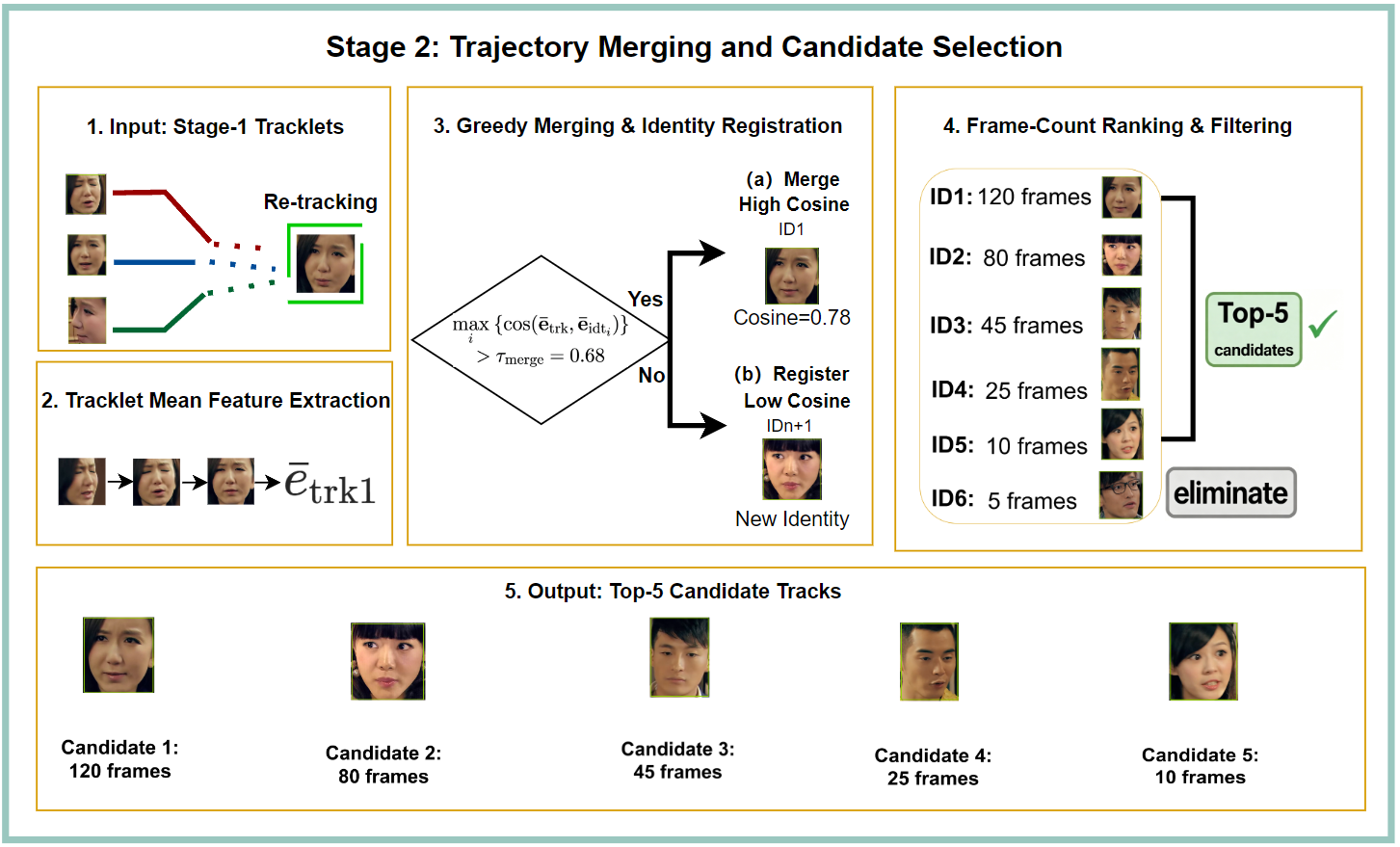}%
}
\caption{Technical pipeline of Stage 2: Trajectory Merging and Candidate Selection.}
\label{f3}
\end{figure}

\textbf{Stage 2: Trajectory Merging and Candidate Selection.}
Due to occlusion or temporary disappearance from the frame, the same speaker may be split into several fragmented sub-trajectories during the tracking stage. For each sub-trajectory, the mean vector of all face embeddings, denoted as $\bar{e}_\text{trk}$, is used as the trajectory-level identity feature. A greedy strategy is then adopted to merge each sub-trajectory into an existing identity cluster: if
\begin{equation}
    \max_{i}\bigl\{
        \cos\!\left(\bar{e}_\text{trk},\,
        \bar{e}_{\text{idt}_i}\right)
    \bigr\} > \tau_\text{merge} = 0.68
    \label{eq:merge_threshold}
\end{equation}
the sub-trajectory is merged into the existing identity with the highest similarity; otherwise, it is registered as a new identity. After trajectory merging, identities are sorted in descending order according to their frame presence, and the top five candidates are selected for subsequent stages. The technical pipeline of this stage is illustrated in Fig.~\ref{f3}.

\par
\begin{figure}[t]
\centering
\makebox[\linewidth][c]{%
    \includegraphics[
        width=0.7\linewidth,
        keepaspectratio
    ]{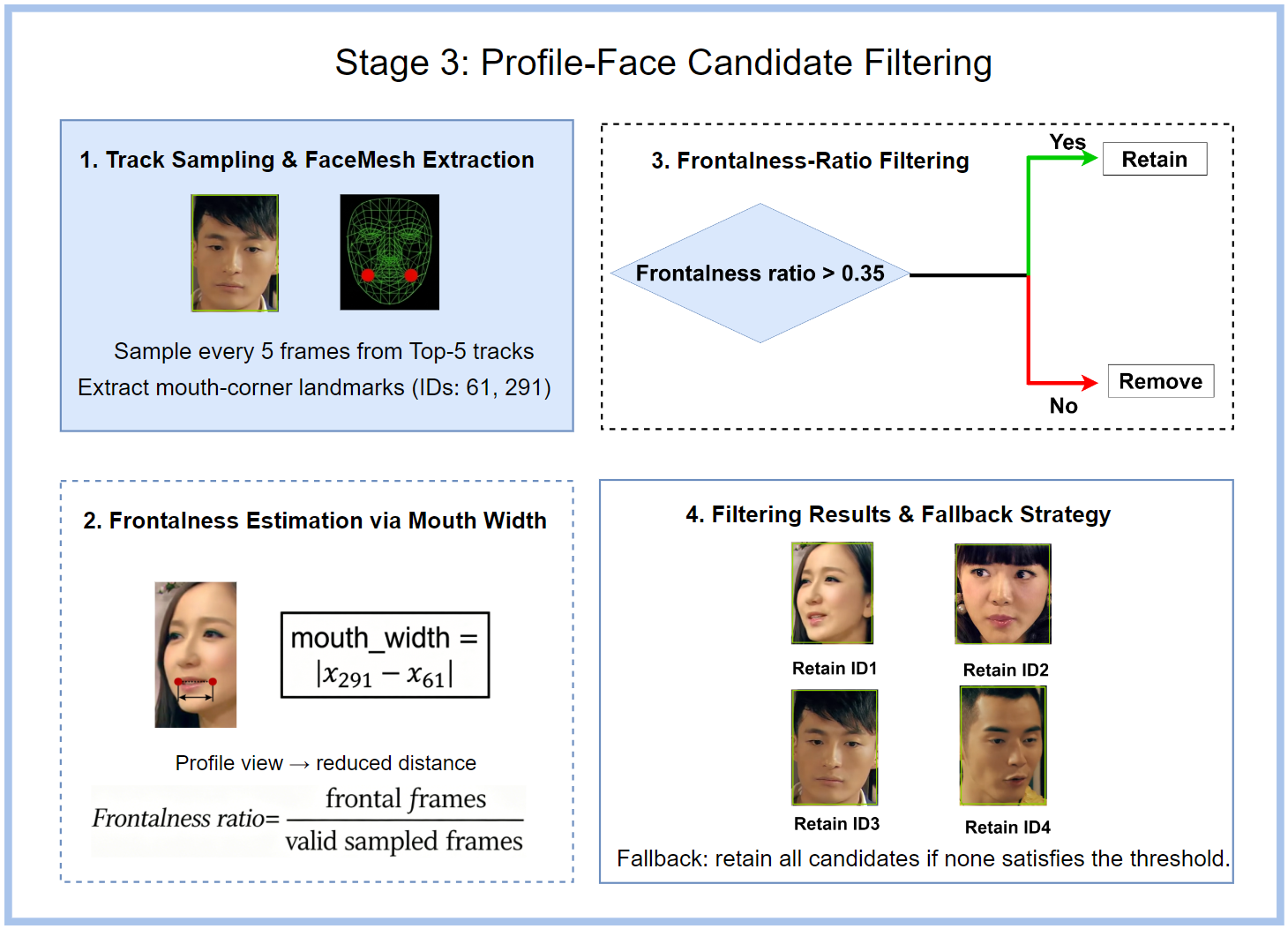}%
}
\caption{Technical pipeline of Stage 3: Profile-Face Candidate Filtering.}
\label{f4}
\end{figure}

\textbf{Stage 3: Profile-Face Candidate Filtering.}
To reduce the interference of profile-face frames in subsequent lip-motion analysis, candidate trajectories are sampled every 5 frames. MediaPipe FaceMesh is first used to extract facial mesh landmarks, and the horizontal distance between the two mouth-corner landmarks, indexed as 61 and 291, is computed as $\text{mouth\_width}=|x_{291}-x_{61}|$. This measurement is used as a proxy indicator of frontal-face degree. Specifically, landmarks 61 and 291 correspond to the mouth-corner region in the MediaPipe FaceMesh topology. In profile-face views, the horizontal span between the mouth corners is significantly reduced in the image projection space, making this indicator sensitive to head pose deviation~\cite{zou2025unsupervised}. A sampled frame is regarded as a frontal-face frame when $\text{mouth\_width} \geq 0.11$. The frontal-face ratio is then defined as the proportion of frontal-face frames among the valid sampled frames of each candidate. Candidates with a frontal-face ratio lower than $35\%$ are discarded to reduce the influence of profile-face trajectories on subsequent lip-motion analysis. If no candidate satisfies this threshold, all candidates are retained as a fallback strategy to prevent pipeline interruption.The technical pipeline of this stage is illustrated in Fig.~\ref{f4}.

\par
\begin{figure}[t]
\centering
\makebox[\linewidth][c]{%
    \includegraphics[
        width=0.7\linewidth,
        keepaspectratio
    ]{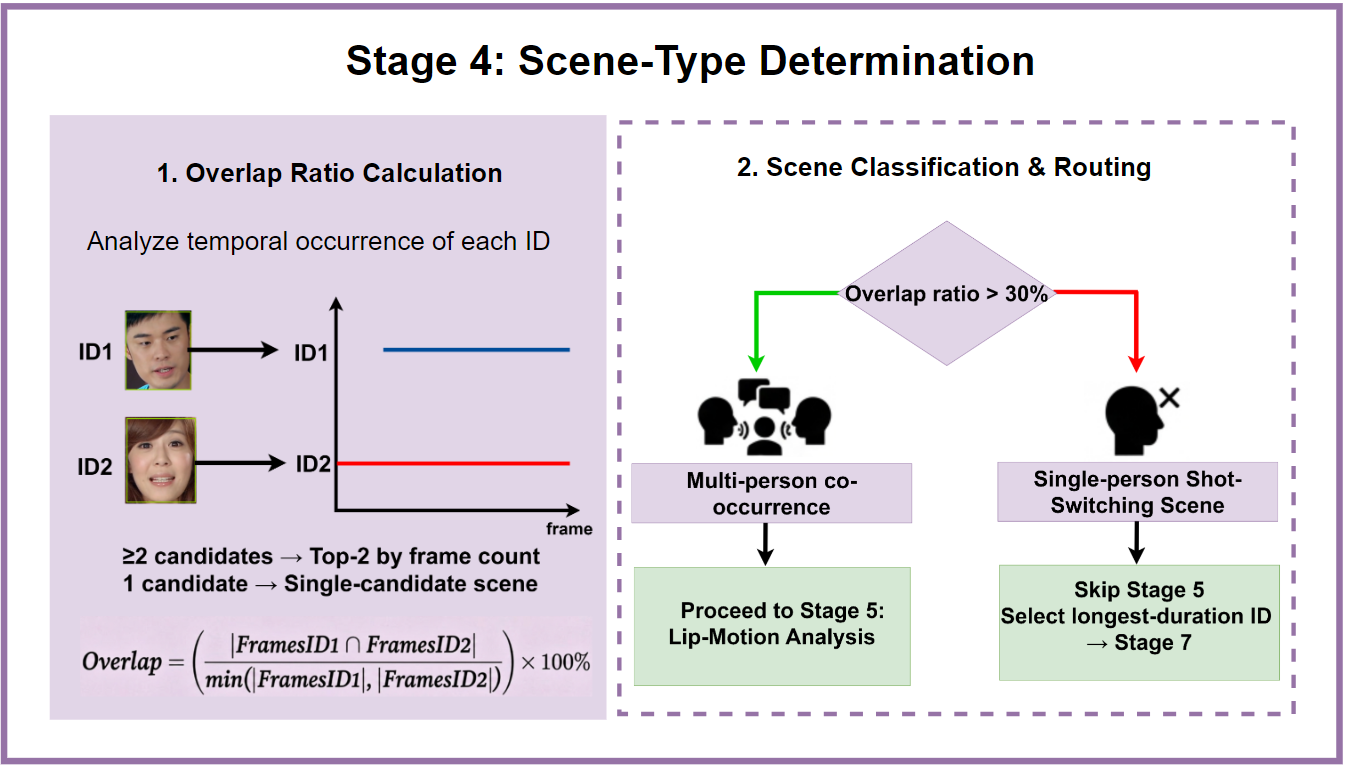}%
}
\caption{Technical pipeline of Stage 4: Scene-Type Determination.}
\label{f5}
\end{figure}

\textbf{Stage 4: Scene-Type Determination.}
When at least two valid candidates are available, the two identities with the largest number of appearing frames are selected as representative candidates, and the overlap ratio of their frame sequences is computed as follows:
\begin{equation}
    \text{Overlap} =
    \frac{
        \bigl|\text{Frames}_{\text{ID}_1}
        \cap \text{Frames}_{\text{ID}_2}\bigr|
    }{
        \min\!\bigl(|\text{Frames}_{\text{ID}_1}|,\,
        |\text{Frames}_{\text{ID}_2}|\bigr)
    }
    \label{eq:overlap}
\end{equation}

\noindent where $\text{Frames}_{\text{ID}_k}$ denotes the set of frame indices in which identity $\text{ID}_k$ appears. The numerator represents the number of co-occurring frames in which the two identities appear simultaneously, while the denominator takes the smaller number of appearing frames between the two identities. This design prevents the overlap ratio from being underestimated when one identity appears for a shorter duration, thereby better reflecting the degree of temporal co-occurrence between the two identities. If only one valid candidate is available, the video is directly treated as a single-candidate scene. If the overlap ratio is no greater than $30\%$, the video is regarded as a single-person shot-switching scene, and active speaker identification is not performed in the subsequent stage. If the overlap ratio is greater than $30\%$, the video is regarded as a multi-person co-occurrence scene and is further processed by active speaker identification in Stage 5. The technical pipeline of this stage is illustrated in Fig.~\ref{f5}.

\par\textbf{Stage 5: Normalized Lip-Motion Analysis.} The previous version used the vertical distance between the upper and lower lips in the normalized FaceMesh coordinate system directly as the speaking score, a measure susceptible to individual differences in mouth shape and to variations in head pose. Two improvements are therefore introduced, both relying exclusively on facial geometric information without any audio or textual cues.

\par\textbf{(1) Lip-Motion Normalization.}
The vertical lip distance is divided by the mouth-corner width to obtain the following dimensionless mouth-opening ratio:
\begin{equation}
    d_i = \frac{|y_{14}-y_{13}|}{|x_{291}-x_{61}|+\varepsilon}
    \label{eq:norm_lip}
\end{equation}
\noindent where the landmark indices follow the MediaPipe FaceMesh topology: $y_{13}$ and $y_{14}$ denote the vertical coordinates of the upper- and lower-lip landmarks, $x_{61}$ and $x_{291}$ the horizontal coordinates of the right and left mouth corners, and $\varepsilon$ is a smoothing term introduced to avoid division by zero. Under image projection, the mouth-corner width and the vertical lip distance generally vary together with changes in head pose, so their ratio compensates for pose-induced scale variation while reducing the influence of individual differences in mouth shape.
\par\textbf{(2) Pose-Aware Soft Weighting.}
Replacing the previous hard-discarding strategy, frame-level samples are weighted continuously according to frontal-face visibility: frontal frames with $\mathit{mouth\_width} \geq 0.110$ receive a weight of $w_i=1.0$, mildly profile frames with $0.077 \leq \mathit{mouth\_width} < 0.110$ receive $w_i=0.5$, and severe-profile frames with $\mathit{mouth\_width} < 0.077$ receive $w_i=0$ and are excluded from the lip-motion statistics. Both thresholds were determined through preliminary experiments: 0.110 corresponds to the typical lower bound of normalized mouth-corner width under near-frontal conditions, whereas values below 0.077 indicate that profile rotation has substantially reduced the reliability of lip-distance estimation~\cite{wan2024precise}. Two statistics are then calculated from the weighted sequence:
\begin{equation}
    \text{CV} = \frac{\sigma_w}{\mu_w},\quad
    \text{DR} = \frac{N_\text{trans}}{\sum_i w_i}
    \label{eq:cv_dr}
\end{equation}

\noindent where the weighted mean $\mu_w$ and weighted standard deviation $\sigma_w$ are defined as:
\begin{equation}
    \mu_w = \frac{\sum_i w_i d_i}{\sum_i w_i},\quad
    \sigma_w =
    \sqrt{
    \frac{\sum_i w_i(d_i-\mu_w)^2}{\sum_i w_i}
    }
    \label{eq:weighted_mean_std}
\end{equation}
\noindent where $N_{\text{trans}}$ is the cumulative number of transitions between mouth-opening states, each state determined by $d_i \geq 0.15$, in the retained normalized mouth-opening sequence, and $\sum_i w_i$ is the total weight, which reduces the influence of differences in video duration. The weighted coefficient of variation CV measures the magnitude of lip motion, whereas the dynamic transition rate DR measures the frequency of transitions between mouth-opening states; together, the two statistics characterize the motion pattern associated with speaking behavior.

To evaluate the two improvements, two variants are compared with all other preprocessing procedures unchanged: \textit{Abs-Hard}, corresponding to the previous version, uses the unnormalized vertical lip distance and hard-filters profile frames, whereas \textit{Norm-Soft}, adopted in this study, uses the mouth-width-normalized mouth-opening ratio and softly weights mildly profile frames according to pose visibility. Of 100 multi-person co-occurrence candidate samples manually annotated from CH-SIMS v2.0S, 81 remained after face tracking, trajectory merging, and profile-face filtering removed samples with large profile angles or insufficient valid frames. Representative results are presented in Table~\ref{t2}.

\begin{table}[htbp]
\centering
\scriptsize
\setlength{\tabcolsep}{2.5pt}
\renewcommand{\arraystretch}{1.15}

\resizebox{0.98\linewidth}{!}{%
\begin{tabular}{@{}l l c c c c c c c c c c@{}}
\hline
Sample
& Candidate
& \multicolumn{2}{c}{Valid frame ratio}
& \multicolumn{3}{c}{Abs-Hard}
& \multicolumn{3}{c}{Norm-Soft}
& \multicolumn{2}{c}{Correct identification} \\
\cline{3-4}
\cline{5-7}
\cline{8-10}
\cline{11-12}

&
& Abs-Hard
& Norm-Soft
& CV
& DR
& Score
& CV
& DR
& Score
& Abs-Hard
& Norm-Soft \\
\hline

Sample 1
& Target speaker
& 100\%
& 100\%
& 0.466
& 0.186
& 0.087
& 0.460
& 0.167
& 0.077
& $\checkmark$
& $\checkmark$ \\

&
Non-target speaker
& 100\%
& 100\%
& 0.900
& 0.000
& 0.000
& 0.736
& 0.000
& 0.000
&
& \\
\hline

Sample 2
& Target speaker
& 100\%
& 100\%
& 0.625
& 0.226
& 0.141
& 0.549
& 0.140
& 0.077
& $\checkmark$
& $\checkmark$ \\

&
Non-target speaker
& 100\%
& 100\%
& 1.631
& 0.000
& 0.000
& 1.364
& 0.000
& 0.000
&
& \\
\hline

Sample 3
& Target speaker
& 100\%
& 100\%
& 0.705
& 0.217
& 0.153
& 0.699
& 0.268
& 0.187
& $\checkmark$
& $\checkmark$ \\

&
Non-target speaker
& 100\%
& 100\%
& 0.891
& 0.000
& 0.000
& 0.650
& 0.000
& 0.000
&
& \\
\hline

Sample 4$^\dagger$
& Target speaker
& 100\%
& 100\%
& 0.590
& 0.205
& 0.121
& 0.545
& 0.128
& 0.070
& $\times$
& $\checkmark$ \\

&
Non-target speaker
& 100\%
& 100\%
& 0.716
& 0.192
& 0.138
& 0.689
& 0.013
& 0.009
&
& \\
\hline

Sample 5$^\ddagger$
& Target speaker
& 54.5\%
& 65.2\%
& 0.000
& 0.000
& 0.000
& 0.355
& 0.034
& 0.012
& $\times$
& $\checkmark$ \\

&
Non-target speaker
& 78.6\%
& 78.6\%
& 0.594
& 0.000
& 0.000
& 0.594
& 0.000
& 0.000
&
& \\
\hline

\multicolumn{2}{l}{Overall identification accuracy}
& \multicolumn{2}{c}{--}
& \multicolumn{3}{c}{70/81 = 86.4\%}
& \multicolumn{3}{c}{73/81 = 90.1\%}
& 86.4\%
& 90.1\% \\
\hline

\end{tabular}%
}

\caption{Ablation comparison of Abs-Hard and Norm-Soft
in the lip-motion analysis module}
\label{t2}

\vspace{2pt}

\begin{minipage}{0.98\linewidth}
\scriptsize
\raggedright
\textit{Note:}
$^\dagger$ denotes a case with sufficient frontal-face frames
but noticeable interference from individual mouth-shape
differences under the unnormalized lip-distance measure.
$^\ddagger$ denotes a pose-constrained case in which the target
speaker has less than 60\% frontal-face frames.
The valid frame ratio is the proportion of a candidate's
appearing frames retained for lip-motion analysis.
Abs-Hard retains frames with
$\mathit{mouth\_width}\geq 0.110$, whereas Norm-Soft
additionally includes mildly profile frames with
$0.077\leq\mathit{mouth\_width}<0.110$
using a weight of 0.5.
Score = CV $\times$ DR.
$\checkmark$ and $\times$ indicate whether the highest-scoring
candidate matches the manual annotation.
\end{minipage}

\end{table}

\begin{figure}[t]
\centering
\makebox[\linewidth][c]{%
    \includegraphics[
        width=1\linewidth,
        keepaspectratio
    ]{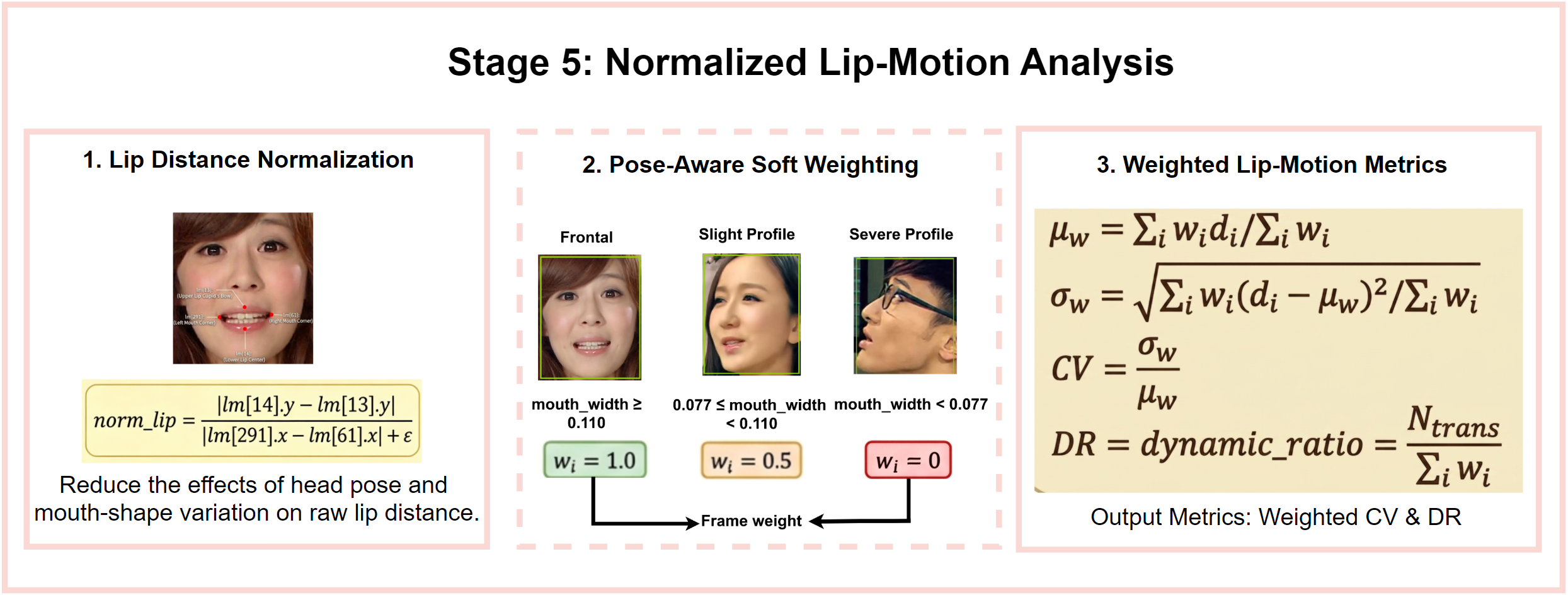}%
}
\caption{Technical pipeline of Stage 5: Normalized Lip-Motion Analysis.}
\label{f6}
\end{figure}

The two improvements serve different purposes. Lip-motion normalization remains neutral when sufficient frontal frames are available: Samples 1--3 produce identical identification results under both variants, while Norm-Soft consistently reduces CV and compresses the variation caused by differences in mouth shape. In Sample 4, it separates two fully frontal candidates whose DR values are excessively close under the unnormalized setting, thereby correcting the misidentification produced by Abs-Hard. Pose-aware soft weighting instead targets pose-constrained scenes: in Sample 5, where only 54.5\% of the target speaker's frames are frontal, hard filtering causes both statistics to degenerate to zero, whereas retaining mildly profile frames raises the proportion of valid frames to 65.2\% and restores the speaking score to a non-zero value. Across all 81 valid multi-person co-occurrence samples, Norm-Soft achieves an identification accuracy of 90.1\%, an improvement of 3.7 percentage points over the 86.4\% achieved by Abs-Hard, validating both normalized lip-motion analysis and pose-aware soft weighting; the technical workflow of this stage is summarized in Fig.~\ref{f6}.

\par
\begin{figure}[t]
\centering
\makebox[\linewidth][c]{%
    \includegraphics[
        width=1\linewidth,
        keepaspectratio
    ]{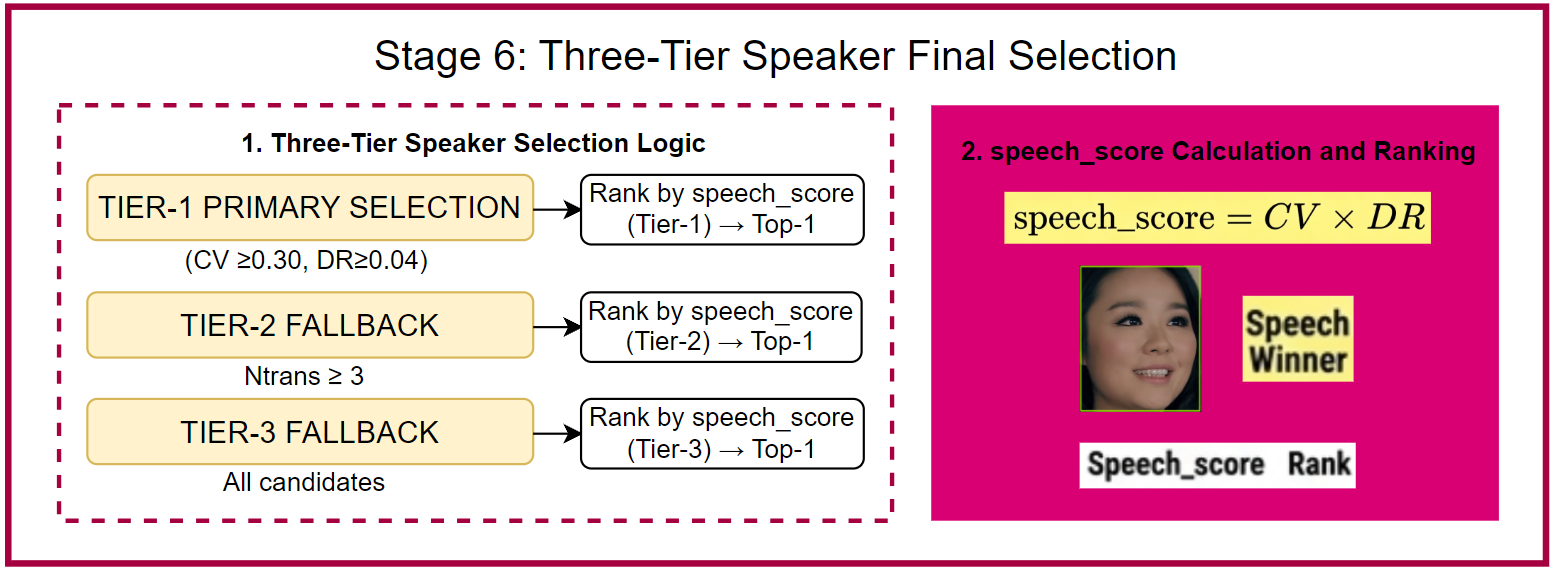}%
}
\caption{Technical pipeline of Stage 6: Three-Tier Speaker Final Selection.}
\label{f7}
\end{figure}

\textbf{Stage 6: Three-Tier Speaker Final Selection.}
For non-multi-person scenes, final selection is not performed and the candidate with the longest frame presence is taken as the target speaker. For multi-person co-occurrence scenes, the final speaking score is defined as $\text{Score}=\text{CV}\times\text{DR}$. This multiplicative form penalizes two types of false positives simultaneously---occasional large mouth openings with high CV but low DR, and continuous minor fluctuations with high DR but low CV---so that a prominent score arises only when both the motion amplitude and the switching frequency are sufficiently significant. Final selection then follows a hierarchical fallback mechanism: \textbf{Tier 1} requires a candidate to satisfy both $\text{CV}\geq0.30$ and $\text{DR}\geq0.04$ and selects the highest-scoring one; if no candidate qualifies, \textbf{Tier 2} requires $N_\text{trans}\geq3$ and again selects the highest-scoring one; if neither tier yields a valid candidate, \textbf{Tier 3} selects the candidate with the highest score among all candidates. This strategy enables strict identification of the active speaker in multi-person co-occurrence scenes while ensuring that the pipeline still produces an output under extreme conditions. The technical pipeline of this stage is illustrated in Fig.~\ref{f7}.

\par
\begin{figure}[t]
\centering
\makebox[\linewidth][c]{%
    \includegraphics[
        width=0.7\linewidth,
        keepaspectratio
    ]{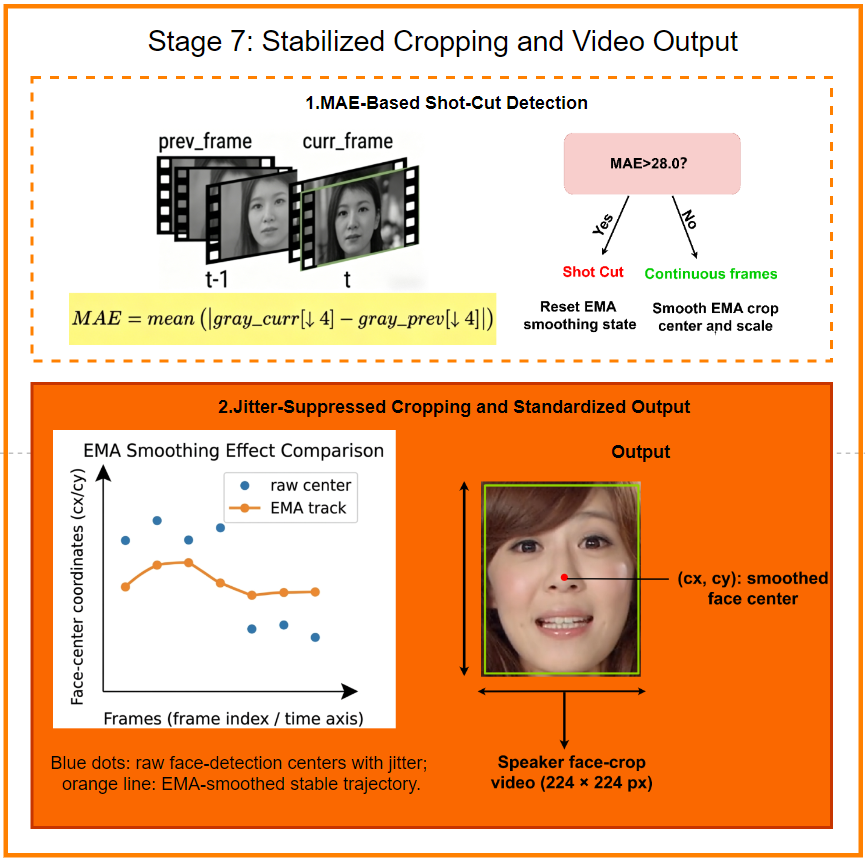}%
}
\caption{Technical pipeline of Stage 7:Stabilized Cropping and Video Output.}
\label{f8}
\end{figure}

\textbf{Stage 7: Stabilized Cropping and Video Output.}
To obtain stable facial videos of the target speaker, both the position and the scale of the cropping window are stabilized in the output stage. Shot transitions are detected from the mean absolute error (MAE) between adjacent grayscale frames downsampled to $1/4$ resolution, with the threshold set to 28.0. The smoothing state is reset immediately at a shot transition, at the first appearance of the target box, or when the cropping scale changes by more than 30\% or the horizontal center shifts by more than 15\% of the frame width, which avoids cross-shot drift and abnormal smoothing. Exponential moving average (EMA, $\alpha=0.1$) smoothing is then applied simultaneously to the cropping center coordinates and the crop side length:

\begin{equation}
\begin{aligned}
    cx_t &= 0.9\,cx_{t-1} + 0.1\,cx_t^{\det},\\
    cy_t &= 0.9\,cy_{t-1} + 0.1\,cy_t^{\det},\\
    s_t  &= 0.9\,s_{t-1}  + 0.1\,s_t^{\det}.
\end{aligned}
\label{eq:ema}
\end{equation}

\noindent where $cx_t$, $cy_t$, and $s_t$ denote the smoothed horizontal center coordinate, vertical center coordinate, and crop side length of the current frame, $cx_{t-1}$, $cy_{t-1}$, and $s_{t-1}$ the corresponding smoothed results of the previous frame, and $cx_t^{\det}$, $cy_t^{\det}$, and $s_t^{\det}$ the original cropping parameters computed from the target face bounding box in the current frame. The historical weight of 0.9 makes the position and scale of the cropping window change more smoothly, thereby effectively suppressing frame-level detection jitter. The original crop side length is given by:

\begin{equation}
    s_t^{\det} = \max(W_{\text{box}},\,H_{\text{box}})
    \times 1.4
    \label{eq:side_length}
\end{equation}

\noindent where $W_{\text{box}}$ and $H_{\text{box}}$ denote the width and height of the target face bounding box, respectively, and the expansion factor of 1.4 preserves an appropriate facial context region. To keep the window inside the frame, the target region is cropped from a black padded canvas constructed around the original frame, resized to $224\times224$, and written to the output video; black frames are inserted for frames in which the target face is absent from the trajectory, maintaining frame-rate continuity and strict temporal alignment between the output video and the original clip. The technical pipeline of this stage is illustrated in Fig.~\ref{f8}.
\par\textbf{Low-Light Recovery Mode.}
For samples in which the main pipeline fails due to low illumination, motion blur, or large-angle profile faces, an additional low-light recovery mode reduces the face detection confidence threshold to 0.20, removes the skin-color filtering constraint, and relaxes the minimum trajectory survival length to 3 frames, so as to capture weak facial signals under extreme visual conditions. All recovered candidate identities are exported through EMA-smoothed cropping and the target speaker is selected manually; samples in which no valid identity can be detected are discarded.

\begin{figure}[t]
\centering
\makebox[\linewidth][c]{%
    \includegraphics[
        width=0.8\linewidth,
        keepaspectratio
    ]{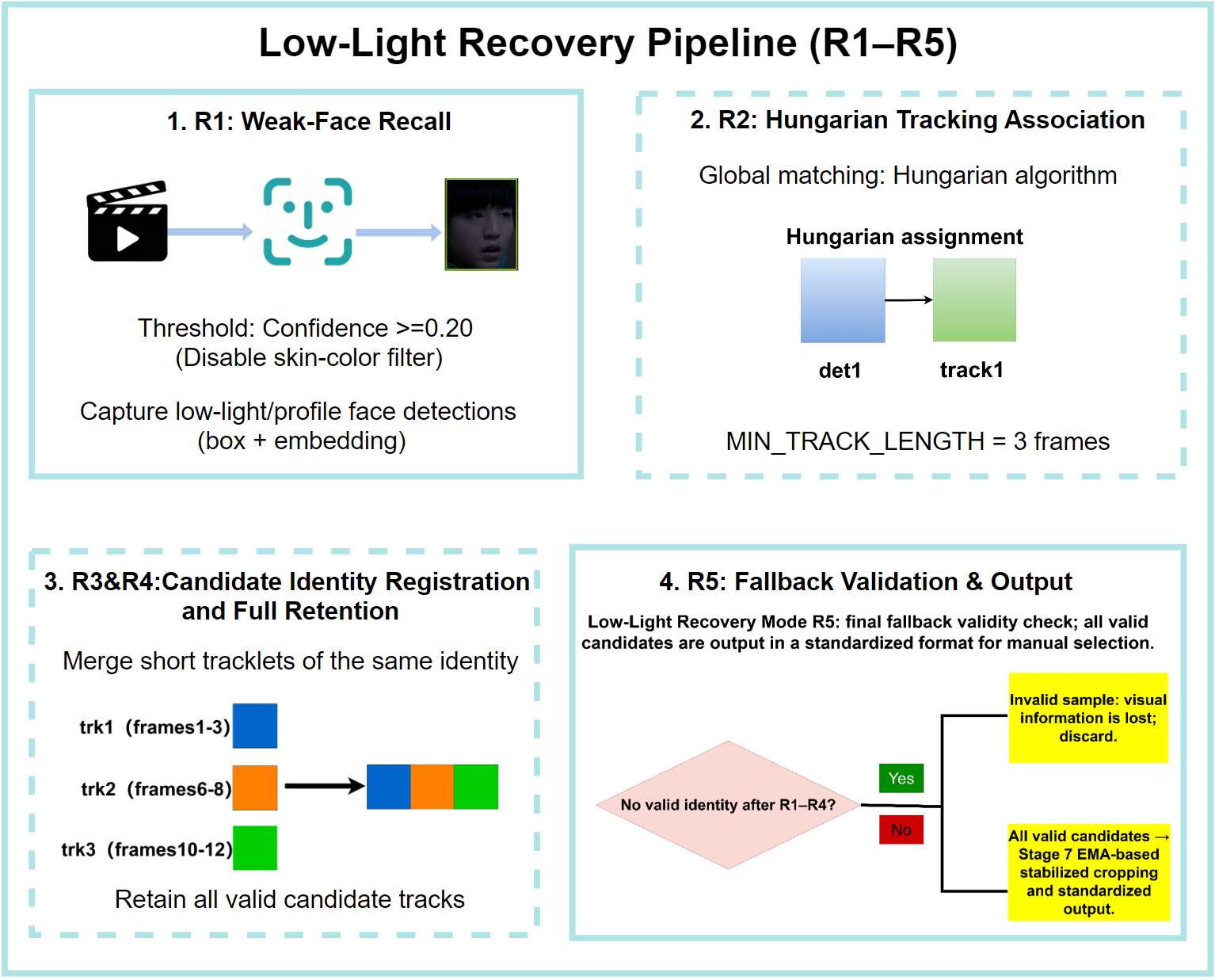}%
}
\caption{Technical pipeline of Low-Light Recovery Mode.}
\label{f9}
\end{figure}

The pipeline is applied to all 4,403 samples of CH-SIMS v2.0S for speaker localization, including neutral samples, which are still processed at this stage and excluded only in the subsequent binary sentiment modeling stage according to the task setting. After automatic localization, low-light recovery, and manual verification, 7 samples are discarded because the speaker remains in a profile-face view throughout the clip, the video is severely blurred, or no valid facial region can be obtained, leaving 4,396 samples with valid localization and cropping results as the valid outputs of the preprocessing stage. The main pipeline achieves an automatic localization accuracy of approximately 95\%, and the remaining approximately 5\% are manually reviewed and corrected before being included in the final preprocessing results. The technical pipeline of this stage is illustrated in Fig.~\ref{f9}.

\section{DS-TAFNet Model Architecture}
\label{sec4}
\subsection{Overall Framework}
\label{subsec4.1}
\begin{figure}[t]
\centering
\makebox[\linewidth][c]{%
    \includegraphics[
        width=1\linewidth,
        keepaspectratio
    ]{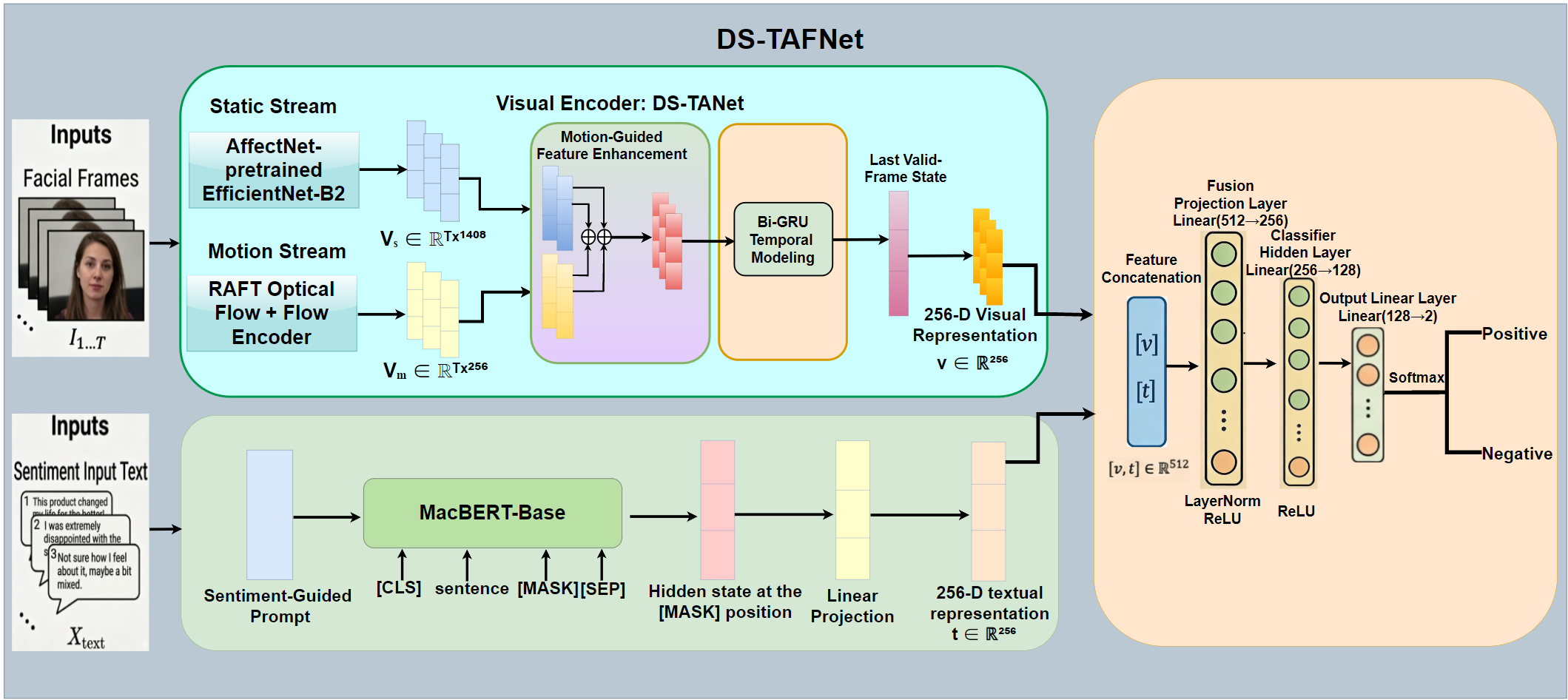}%
}
\caption{Overall architecture of DS-TAFNet}
\label{f10}
\end{figure}

DS-TAFNet consists of a dual-stream temporal visual encoder (DS-TANet), a text encoder, and a multimodal fusion classifier. DS-TANet processes the preprocessed facial-frame sequence through a static branch, which extracts facial appearance features, and a dynamic branch, which uses RAFT to estimate optical flow between adjacent frames and encode facial motion; a motion-guided attention mechanism then enhances the static representation with the motion features, and a Bi-GRU models temporal dependencies to produce a 256-dimensional visual representation. The text encoder uses MacBERT-Base with a sentiment-guided prompt and linearly projects the hidden state at the $\texttt{[MASK]}$ position into a 256-dimensional textual representation. The multimodal fusion classifier concatenates the two representations and passes them through a fusion projection layer and a classification layer to predict sentiment polarity. The overall architecture is shown in Fig.~\ref{f10}.

\subsection{Dual-Stream Temporal Visual Encoder (DS-TANet)}
\label{subsec4.2}

DS-TANet takes a uniformly sampled facial-frame sequence generated by the NAPS pipeline described in Section 3 as input. Each video clip contains $T=14$ frames resized to $224\times224$. The encoder adopts a dual-stream architecture to model frame-level static appearance features and inter-frame dynamic motion features, respectively. The resulting features are enhanced through a motion-guided mechanism and subsequently processed by a Bi-GRU for global temporal modeling.

\textbf{Static stream.}
The static stream employs EfficientNetB2 pretrained on AffectNet~\cite{mollahosseini2019affectnet}. Each frame is independently encoded into a frame-level appearance representation $\mathbf{f}_s^t\in\mathbb{R}^{1408}$ ($t=1,\ldots,T$), forming the static feature sequence $\mathbf{F}_s\in\mathbb{R}^{T\times1408}$. Pretraining on AffectNet provides the backbone with discriminative capability for facial-expression analysis, thereby facilitating the extraction of static emotional cues within individual frames.

\textbf{Dynamic stream.}
The dynamic stream employs a frozen RAFT network to estimate dense optical flow $(u,v)$ between adjacent frames. Each optical-flow map is compressed using a three-layer convolutional encoder consisting of stride-2 convolution, batch normalization, and ReLU activation, followed by adaptive average pooling and flattening. This process produces a frame-level motion representation $\mathbf{f}_d^t\in\mathbb{R}^{256}$. Because the first frame has no preceding frame for optical-flow estimation, a zero vector is inserted at the beginning of the sequence to maintain temporal alignment. The complete dynamic feature sequence is denoted by $\mathbf{F}_d\in\mathbb{R}^{T\times256}$.

\textbf{Motion-guided feature enhancement.}
Rather than directly concatenating the two streams, DS-TANet uses the dynamic feature as a modulation signal to enhance the static feature in a channel-wise manner through multiplicative residual modulation:

\begin{equation}
    \mathbf{m}^t = \sigma\!\left(
        \mathbf{W}_2\,\operatorname{ReLU}\!\left(
            \mathbf{W}_1\,\mathbf{f}_d^t
        \right)
    \right),\quad
    \mathbf{f}^t = \mathbf{f}_s^t \odot
    \left(1 + \mathbf{m}^t\right)
    \label{eq:motion_attn}
\end{equation}

\noindent where $\mathbf{W}_1\in\mathbb{R}^{1408\times256}$ projects the 256-dimensional motion feature into the same 1408-dimensional space as the static feature, and $\mathbf{W}_2\in\mathbb{R}^{1408\times1408}$ denotes the second linear transformation. $\sigma(\cdot)$ denotes the Sigmoid activation function. $\mathbf{m}^t\in[0,1]^{1408}$ is a channel-wise modulation mask that indicates the degree to which each static feature channel is enhanced by motion information at time step $t$, and $\odot$ denotes element-wise multiplication. The residual term $(1+\mathbf{m}^t)$ preserves the baseline static representation when the motion signal is weak, thereby reducing interference from noisy dynamic features.

\textbf{Bi-GRU temporal modeling.}
Enhanced features corresponding to invalid black frames are set to zero using a valid-frame mask. The sequence ${\mathbf{f}^t}{t=1}^{T}$ is then processed by a single-layer bidirectional GRU~\cite{cho2014gru} with 128 hidden units in each direction. Concatenating the two directional states yields a 256-dimensional representation $\mathbf{h}^t$ for each frame. The state at the last valid frame is selected as the video-level representation:
\begin{equation}
    \mathbf{H} = \operatorname{Bi\text{-}GRU}
    \!\left(\{\mathbf{f}^t\}_{t=1}^{T}\right),\quad
    \mathbf{v} = \mathbf{h}^{T^*}
    \label{eq:bi_gru}
\end{equation}

\noindent where $\mathbf{h}^t\in\mathbb{R}^{256}$ concatenates the 128-dimensional forward and backward states at frame $t$, $T^{*}$ is the index of the last valid frame, and $\mathbf{v}\in\mathbb{R}^{256}$ is the video-level visual representation. $T^{*}$ is determined from the valid-frame mask, preventing the model from selecting a fixed final time step that may correspond to an invalid or black frame.

\subsection{Text Encoder}

The text encoder uses MacBERT-Base with a fixed sentiment-guided prompt~\cite{liu2023pretrain}: ``Overall, the sentiment tendency of this text is \texttt{[MASK]}.'' The $\texttt{[MASK]}$ token acts as a sentiment-aware semantic anchor that aggregates polarity-related contextual information.

The final-layer hidden state at the $\texttt{[MASK]}$ position is projected to the same dimensionality as the visual representation:
\begin{equation}
    \mathbf{t} = \operatorname{Proj}\!\left(
        \operatorname{MacBERT}(\mathbf{x})_{\texttt{[MASK]}}
    \right),\quad \mathbf{t}\in\mathbb{R}^{256}
    \label{eq:text_enc}
\end{equation}

\noindent where $\mathbf{x}$ is the prompt-enhanced text sequence, and $\operatorname{MacBERT}(\cdot)_{\texttt{[MASK]}}$ denotes the 768-dimensional final-layer hidden state at the $\texttt{[MASK]}$ position. $\operatorname{Proj}$ comprises a linear layer, LayerNorm, GELU, and Dropout, and maps the hidden state to a 256-dimensional textual representation. Training follows two stages. First, MacBERT is frozen while the projection, fusion, and classification modules are trained. Second, its top two layers are unfrozen and jointly fine-tuned with a smaller learning rate, enabling task adaptation while preserving pretrained semantics.
\subsection{Multimodal Fusion and Classification}
\label{subsec4.4}
Within the DS-TAFNet framework, this paper systematically compares two fusion strategies to analyze how data scale constrains the selection of cross-modal fusion mechanisms. Both strategies take the visual representation $\mathbf{v}\in\mathbb{R}^{256}$ and the textual representation $\mathbf{t}\in\mathbb{R}^{256}$ as input, and perform fusion after dimensional alignment. They differ in their fusion structures and cross-modal interaction mechanisms, as described below.

\textbf{Concat Fusion (the finally adopted lightweight fusion strategy).}
Concat Fusion directly concatenates the visual and textual representations, feeds the concatenated vector into a fusion head, and then uses a two-layer linear classifier to produce the sentiment polarity prediction:
\begin{align}
    \mathbf{z} &= \operatorname{ReLU}\!\left(
        \operatorname{LayerNorm}\!\left(
            \mathbf{W}_f\,[\mathbf{v};\,\mathbf{t}]
            + \mathbf{b}_f
        \right)
    \right) \label{eq:concat_z} \\
    \hat{y} &= \operatorname{Softmax}\!\left(
        \mathbf{W}_2\,\operatorname{ReLU}\!\left(
            \mathbf{W}_1\,\mathbf{z}
        \right)
    \right) \label{eq:concat_fusion}
\end{align}

\noindent where $[\mathbf{v};\mathbf{t}]\in\mathbb{R}^{512}$ denotes the concatenated visual-textual representation. $\mathbf{W}_f\in\mathbb{R}^{256\times512}$ and $\mathbf{b}_f\in\mathbb{R}^{256}$ denote the weight matrix and bias term of the linear transformation in the fusion head, respectively. They map the concatenated vector to a 256-dimensional fused representation, followed by LayerNorm and ReLU activation to obtain the intermediate representation $\mathbf{z}$. $\mathbf{W}_1\in\mathbb{R}^{128\times256}$ and $\mathbf{W}_2\in\mathbb{R}^{2\times128}$ denote the weight matrices of the two-layer classifier. $\hat{y}$ denotes the class probability after Softmax transformation. For simplicity, the bias terms in the classifier and implementation details such as Dropout are omitted in the equations. In the actual implementation, Dropout is applied in both the fusion head and the main classifier to reduce the risk of overfitting. This strategy has approximately $0.4$M trainable parameters during fusion training and is relatively lightweight. To improve generalization stability under small-data conditions, Concat Fusion further introduces Modal Dropout ($p=0.15$) and auxiliary classification heads (Auxiliary Loss, with a weight of $0.3$). The former independently zeros out each modality feature with a probability of 15\% during training, forcing the fusion head to remain effective under single-modality conditions. The latter attaches an auxiliary classification head to the visual and textual features, respectively, imposing additional supervision on the two unimodal representations to enhance their independent discriminative capability and reduce excessive reliance on either modality during fusion.

\textbf{Cross-Modal Attention (comparison strategy).}
Cross-Modal Attention adopts a bidirectional cross-modal attention mechanism to enhance the interaction between visual and textual representations~\cite{vaswani2017attention}. Specifically, the visual representation $\mathbf{v}$ is used as the Query, while the textual representation $\mathbf{t}$ is used as the Key and Value for multi-head attention. The symmetric direction is also performed, where $\mathbf{t}$ queries $\mathbf{v}$. Each attention branch contains two sublayers, including a residual connection with layer normalization and a feed-forward network (FFN). The computation is formulated as follows:
\begin{align}
    \mathbf{v}_1 &= \operatorname{LN}\!\left(
        \mathbf{v} +
        \operatorname{MHA}(\mathbf{v},\mathbf{t},\mathbf{t})
    \right), \label{eq:crossattn_v1} \\
    \mathbf{v}' &= \operatorname{LN}\!\left(
        \mathbf{v}_1 + \operatorname{FFN}(\mathbf{v}_1)
    \right), \label{eq:crossattn_v2} \\
    \mathbf{t}_1 &= \operatorname{LN}\!\left(
        \mathbf{t} +
        \operatorname{MHA}(\mathbf{t},\mathbf{v}_{\text{orig}},\mathbf{v}_{\text{orig}})
    \right), \label{eq:crossattn_t1} \\
    \mathbf{t}' &= \operatorname{LN}\!\left(
        \mathbf{t}_1 + \operatorname{FFN}(\mathbf{t}_1)
    \right). \label{eq:crossattn_t2}
\end{align}

\noindent where $\operatorname{MHA}(Q,K,V)$ denotes a 4-head multi-head attention function, $\operatorname{LN}(\cdot)$ denotes layer normalization, and $\operatorname{FFN}$ denotes a two-layer feed-forward network with a hidden dimension of $2\times256=512$. $\mathbf{v}_1$ denotes the intermediate visual representation obtained by enhancing $\mathbf{v}$ with the textual representation $\mathbf{t}$ as the Key and Value, followed by residual connection and layer normalization. $\mathbf{t}_1$ denotes the intermediate textual representation obtained by enhancing $\mathbf{t}$ with the original visual representation $\mathbf{v}_{\text{orig}}$ as the Key and Value, followed by residual connection and layer normalization. Subsequently, $\mathbf{v}_1$ and $\mathbf{t}_1$ are separately processed by the feed-forward network, residual connection, and layer normalization to obtain the final enhanced visual representation $\mathbf{v}'$ and textual representation $\mathbf{t}'$. In Eq.~\eqref{eq:crossattn_t1}, $\mathbf{v}_{\text{orig}}$ denotes the original visual representation before attention enhancement, which is distinguished from the updated $\mathbf{v}'$ in Eqs.~\eqref{eq:crossattn_v1}--\eqref{eq:crossattn_v2}. Both attention directions use the original representation of the other modality as the Key and Value, forming a symmetric cross-modal querying design. The enhanced visual representation $\mathbf{v}'\in\mathbb{R}^{256}$ and textual representation $\mathbf{t}'\in\mathbb{R}^{256}$ are then concatenated, projected, and fed into a classification head with the same structure as that used in Concat Fusion. This strategy mainly implements visual-textual feature interaction through cross-modal attention layers and feed-forward networks, without additionally introducing Modal Dropout or auxiliary classification heads. It has approximately $1.4$M trainable parameters during fusion training. The experimental results and performance comparison of the two fusion strategies are further analyzed in Section 5.
\section{Experimental Design and Result Analysis}
\label{sec5}

\subsection{Experimental Setup}
\label{subsec5.1}

\subsubsection{Dataset and Split}
\label{subsubsec5.1.1}

Experiments use the CH-SIMS v2.0S supervised subset under the official split, with neutral samples removed as described in Section 3.1, leaving 3,866 samples (Table 1). Because the training-set positive-to-negative ratio is approximately 1:1.20, balanced sampling and Focal Loss are used to reduce the influence of class imbalance.

\subsubsection{Evaluation Metrics}
\label{subsubsec5.1.2}

Acc2 and Macro F1 are used as evaluation metrics. The positive-to-negative ratios in the training, validation, and test sets are approximately 1:1.20, 1:1.41, and 1:1.15, respectively. Under this class imbalance, accuracy may overestimate performance on the majority class.

Macro F1 calculates the F1 score for each class independently and then averages them with equal weights, making it less sensitive to class frequency. It is therefore used as the primary evaluation metric, while Acc2 is reported as a supplementary metric.

\subsubsection{Training Configuration}
\label{subsubsec5.1.3}

The modality encoders are trained independently before fusion. DS-TANet and all visual ablation variants use AdamW with $\text{lr}=3\times10^{-5}$ and weight decay $5\times10^{-3}$, together with Focal Loss~\cite{lin2017focal} using $\alpha=0.75$ and $\gamma=2.5$. Linear warm-up and cosine annealing are applied. The batch size is 32 and early-stopping patience is 3. All visual variants use identical hyperparameters, with data augmentation, balanced sampling, and Focal Loss applied consistently.

MacBERT-Base uses AdamW with a top-layer learning rate of $1\times10^{-5}$, weight decay of 0.1, and layer-wise learning-rate decay of 0.85. The embedding layer and bottom four Transformer layers are frozen, while the top eight layers are trainable. FGM~\cite{goodfellow2015explaining} and stochastic weight averaging are used to improve robustness. The batch size is 32 and early-stopping patience is 6.

DS-TAFNet is trained in two stages. First, both unimodal backbones are frozen and the fusion parameters are trained for 8 epochs with $\text{lr}=3\times10^{-4}$. Second, the visual encoder remains frozen, while the top two MacBERT layers and the fusion module are jointly fine-tuned for 6 epochs. Their learning rates are $5\times10^{-6}$ and $5\times10^{-5}$, respectively, with a batch size of 16. Focal Loss with label smoothing is used and early-stopping patience is 5. This strategy reduces overfitting while preserving stable unimodal representations.

\subsubsection{Implementation Details}
\label{subsubsec5.1.4}

All experiments are implemented in PyTorch and run on an NVIDIA RTX 4080 GPU. EfficientNetB2 is initialized with AffectNet-pretrained weights, RAFT remains frozen throughout visual and fusion training, and MacBERT-Base uses pretrained weights. Balanced sampling is used in both visual and multimodal training.

\subsection{Comparison with Existing Methods}
\label{subsec5.2}

\subsubsection{Comparison with Baselines Reported in the Original Dataset Paper}
\label{subsubsec5.2.1}

Table~\ref{t3} compares the proposed method with all baselines from the original CH-SIMS v2.0 paper in terms of Acc2. While baselines use trimodal features(T+A+V), our method relies solely on text and vision (T+V) via Concat Fusion; F1 is omitted due to inconsistent evaluation criteria across studies.
Table~\ref{t4} presents the confusion matrix for evaluation transparency: on the 911-sample test set, the negative class achieves Precision 91.6\% / Recall 84.4\%, and the positive class achieves Precision 83.5\% / Recall 91.0\%,
yielding overall Acc2 87.49\% and Macro F1 87.48\%.
\begin{table}[h]
\centering
\small
\setlength{\tabcolsep}{8pt}
\begin{tabular}{l c c c}
\hline
Model & Modality & Additional Data & Acc2 \\
\hline
LF\_DNN ~\cite{williams2018}      & T+A+V & None & 73.95\% \\
TFN ~\cite{zadeh2017}             & T+A+V & None & 76.51\% \\
LMF ~\cite{liu2018}               & T+A+V & None & 77.05\% \\
MFN ~\cite{zadeh2018mfn}          & T+A+V & None & 75.27\% \\
Graph\_MFN ~\cite{zadeh2018mosei} & T+A+V & None & 73.98\% \\
MulT ~\cite{tsai2019multimodal}   & T+A+V & None & 79.50\% \\
Bert\_MAG ~\cite{rahman2020}      & T+A+V & None & 79.79\% \\
MISA ~\cite{hazarika2020}         & T+A+V & None & 80.53\% \\
MMIM ~\cite{han2021}              & T+A+V & None & 80.95\% \\
Self\_MM ~\cite{yu2021}           & T+A+V & None & 79.01\% \\
MLF\_DNN ~\cite{liu2022chsimsv2}  & T+A+V & None & 78.40\% \\
MTFN ~\cite{liu2022chsimsv2}      & T+A+V & None & 80.26\% \\
MLMF ~\cite{liu2022chsimsv2}      & T+A+V & None & 79.92\% \\
AV-MC ~\cite{liu2022chsimsv2}     & T+A+V & None & 82.50\% \\
AV-MC(Semi) ~\cite{liu2022chsimsv2}
& T+A+V & +10,161 unlabeled samples & 83.46\% \\
\hline
\textbf{Ours (DS-TAFNet)}
& \textbf{T+V}
& \textbf{None}
& \textbf{87.49\%} \\
\hline
\end{tabular}
\caption{Acc2 comparison with baselines reported in the original CH-SIMS v2.0 paper}
\label{t3}
\end{table}

\begin{table}[h]
\centering
\small
\begin{tabular}{l c c}
\hline
 & \textbf{Predicted: Negative} & \textbf{Predicted: Positive} \\
\hline
\textbf{Actual: Negative} (488) & 412 (TN) & 76 (FP) \\
\textbf{Actual: Positive} (423) & 38 (FN) & 385 (TP) \\
\hline
\end{tabular}
\caption{Confusion matrix of the proposed method on the test set using Concat Fusion}
\label{t4}
\end{table}

The proposed method achieves an Acc2 of 87.49\%, outperforming all baseline models listed in Table~\ref{t3}. Compared with the highest semi-supervised result, AV-MC(Semi) (83.46\%), and the best fully supervised baseline, AV-MC (82.50\%), the proposed method improves Acc2 by 4.03 and 4.99 percentage points, respectively.

The exclusion of the acoustic modality is motivated by the empirical findings reported in the original paper. Under unimodal-label supervision and evaluation, the default acoustic features achieve an Acc2 of only 57.16\%, substantially lower than the 78.72\% obtained by both the textual and visual modalities. This suggests that the default acoustic representation has relatively limited discriminative capability under the original experimental setting. Therefore, this study focuses on text--visual bimodal modeling by improving visual preprocessing and temporal encoding. Moreover, AV-MC(Semi) uses 10,161 additional unlabeled videos, approximately 4.2 times the size of the training set used in this study, whereas the proposed method achieves superior performance without using any additional data.

\subsubsection{Comparison with Recent Related Methods}
\label{subsubsec5.2.2}
Table~\ref{t5} compares the proposed method with recent methods on CH-SIMS v2.0S. Results marked with $^*$ are taken from the open-source reproductions reported by Zhong et al.~\cite{zhong2025cmc}. All methods use Acc2 based on the overall sentiment label of each video segment as the evaluation metric. This study follows the official dataset split, while additionally removing neutral samples and samples that cannot provide valid facial sequences after the NAPS visual quality-control procedure. For the retained samples, the label definition and Acc2 calculation remain consistent with those used in the compared methods. The proposed method outperforms CMC and MC-Teacher by 3.06 and 3.16 percentage points, respectively. Zhong et al.~\cite{zhong2025cmc} reported an acoustic unimodal Acc2 of 61.12\%, which is lower than the visual unimodal Acc2 of 76.31\%. This result supports the research choice of prioritizing the improvement of textual and visual representation quality.

NAPS improves visual input quality through face detection, trajectory association, speaker selection, and invalid-frame filtering. DS-TANet uses EfficientNetB2, RAFT, and Bi-GRU to model appearance, motion, and temporal information, respectively, while prompted MacBERT is used for textual modeling. Without introducing the acoustic modality or any additional data, the proposed method ultimately achieves an Acc2 of 87.49\% and a Macro F1 of 87.48\%.

\begin{table}[h]
\centering
\begin{tabular}{l c c c}
\hline
Model & Modality & Additional Data & Acc2 \\
\hline
\multicolumn{4}{l}{Trimodal methods (T+A+V)} \\
\hline
AV-MC ~\cite{liu2022chsimsv2}
& T+A+V & None & 82.50\% \\

EMT-DLFR ~\cite{sun2024emt}$^*$
& T+A+V & None & 78.63\% \\

TriagedMSA ~\cite{luo2025triaged}
& T+A+V & None & 83.75\% \\

MC-Teacher ~\cite{yuan2024mcteacher}
& T+A+V & +10,161 unlabeled samples & 84.33\% \\

CMC ~\cite{zhong2025cmc}
& T+A+V & None & 84.43\% \\
\hline
\multicolumn{4}{l}{Bimodal methods (T+V)} \\
\hline
\textbf{Ours (DS-TAFNet)}
& \textbf{T+V}
& \textbf{None}
& \textbf{87.49\%} \\
\hline
\end{tabular}
\caption{Acc2 comparison with recent methods on CH-SIMS v2.0S}
\label{t5}

\begin{minipage}{\linewidth}
\footnotesize
$^*$ indicates an open-source reproduction result reported by Zhong et al.~\cite{zhong2025cmc}.
MC-Teacher uses 10,161 additional unlabeled samples.
\end{minipage}
\end{table}

\subsection{Ablation Study of the Visual Encoder}
\label{subsubsec5.3}
To assess the contribution of each component in DS-TANet, we conduct five ablation experiments (Table~\ref{t6}): starting from an EfficientNetB2 static-stream baseline (Exp1), we add the RAFT dynamic stream and compare Concat against motion-guided attention fusion, each with and without Bi-GRU temporal modeling.
\begin{table}[h]
\centering
\small
\setlength{\tabcolsep}{4pt} 
\begin{tabular}{lcccccc}
\hline
ID & \makecell{Static \\ Stream} & \makecell{Dynamic \\ Stream} & Fusion & \makecell{Temporal \\ Modeling} & Acc2 & Macro F1 \\
\hline
Exp1 & $\checkmark$ & --- & --- & --- & 0.8101 & 0.8098 \\
Exp2 & $\checkmark$ & $\checkmark$ & Concat & --- & 0.8189 & 0.8187 \\
Exp3 & $\checkmark$ & $\checkmark$ & Concat & Bi-GRU & 0.8222 & 0.8212 \\
Exp4 & $\checkmark$ & $\checkmark$ & Attention & --- & 0.8101 & 0.8100 \\
Exp5 & $\checkmark$ & $\checkmark$ & Attention & Bi-GRU & \textbf{0.8277} & \textbf{0.8258} \\
\hline
\end{tabular}
\caption{Ablation results of the visual encoder}
\label{t6}
\end{table}

The full model (Exp5) performs best, improving Acc2 and Macro F1 by 1.76 and 1.60 percentage points over the static baseline. Of this, the RAFT dynamic stream alone (Exp1$\to$Exp2) accounts for 0.89 points of F1, confirming that optical-flow motion cues benefit visual sentiment recognition.

The benefit of attention fusion is strongly condition-dependent: without temporal modeling it falls below simple concatenation (Exp4 vs.\ Exp2, $-0.87$ points), but with Bi-GRU it overtakes it (Exp5 vs.\ Exp3, $+0.46$ points). Consistently, the gain from Bi-GRU is far larger along the attention path ($+1.58$) than along the Concat path ($+0.25$). Channel modulation from motion-guided attention therefore appears to require temporal context to take effect, indicating a complementary relationship between the two mechanisms.

\subsection{Unimodal Baseline Performance}
\label{subsec5.4}

The text encoder, MacBERT-Base with a sentiment-guided prompt, and the visual encoders, including the Exp1 static baseline and the complete Exp5 DS-TANet, are trained independently. The results are shown in Table~\ref{t7}.

The results exhibit a two-level progression. Exp1, which uses only the EfficientNetB2 static stream, achieves a Macro F1 of 80.98\%, nearly matching the text baseline of 80.55\%, with a difference of only 0.43 percentage points. This demonstrates that NAPS preprocessing alone can produce high-quality facial sequences for static visual feature extraction. After incorporating the RAFT dynamic stream, motion-guided attention fusion, and Bi-GRU temporal modeling, Exp5 achieves a Macro F1 of 82.58\%, representing an improvement of 1.60 percentage points over Exp1. It also clearly outperforms the text baseline, validating the incremental contribution of the architectural design.

These results confirm that the improvement in visual performance stems from the synergy between preprocessing and architectural design. Since CH-SIMS v2.0S contains many samples that require nonverbal cues for sentiment judgment, the visual modality's comparable or superior discriminative performance relative to the textual modality highlights the practical value of improving visual input quality and temporal representation.
\begin{table}[h]
\centering
\begin{tabular}{l c c c}
\hline
Modality & Model & Acc2 & Macro F1 \\
\hline
Text
& MacBERT-Base$^\dagger$
& 0.8079
& 0.8055 \\
\hline
Vision
& Exp1 (EfficientNetB2, static baseline)$^\ddagger$
& 0.8101
& 0.8098 \\
Vision
& Exp5 (DS-TANet, complete architecture)$^\S$
& \textbf{0.8277}
& \textbf{0.8258} \\
\hline
\end{tabular}
\caption{Comparison of unimodal baselines}
\label{t7}

\begin{minipage}{\linewidth}
\footnotesize
$^\dagger$ MacBERT-Base fine-tuned with a sentiment-guided prompt.\\
$^\ddagger$ NAPS-preprocessed visual input using only the EfficientNetB2 static stream, without the dynamic stream or temporal modeling.\\
$^\S$ NAPS-preprocessed visual input using the EfficientNetB2 static stream and RAFT dynamic stream, combined with motion-guided attention fusion and Bi-GRU temporal modeling.
\end{minipage}
\end{table}

\subsection{Comparison of Fusion Strategies}
\label{subsec5.5}

Using Exp5 and MacBERT-Base as the visual and textual backbones, respectively, this study compares Concat Fusion, with approximately $0.4$M parameters, and Cross-Modal Attention, with approximately $1.4$M parameters. The results are shown in Table~\ref{t8}.

Cross-Modal Attention achieves a slightly higher test-set Macro F1 than Concat Fusion, increasing the score from 0.8748 to 0.8814, corresponding to an improvement of 0.66 percentage points. However, its parameter count is approximately 3.5 times that of Concat Fusion, which introduces greater validation instability under the constraint of approximately 2.4k training samples~\cite{zhang2026multimodal}. In contrast, Concat Fusion has lower model complexity, a more stable training process, and stronger generalization robustness. Therefore, considering model complexity, training stability, and test-set performance, Concat Fusion is adopted as the recommended fusion strategy for DS-TAFNet.

\begin{table}[h]
\centering
\begin{tabular}{l c c c}
\hline
Fusion Strategy & Fusion Parameters & Acc2 & Macro F1 \\
\hline
Concat Fusion
& $\sim$0.4M
& 0.8749
& 0.8748 \\
Cross-Modal Attention
& $\sim$1.4M
& 0.8814
& 0.8814 \\
\hline
\end{tabular}
\caption{Comparison of fusion strategies}
\label{t8}
\end{table}

\section{Conclusion}
\label{sec6}

To address insufficient visual representation quality in multimodal sentiment
analysis, this paper introduces systematic improvements at three levels---visual preprocessing, temporal visual modeling, and multimodal fusion---evaluated on the supervised subset of CH-SIMS v2.0S.

At the visual preprocessing level, NAPS is a seven-stage automated pipeline
comprising face tracking, identity association, normalized lip-motion analysis,
and pose-aware soft weighting, achieving an automatic localization accuracy of
approximately 95\%. The EfficientNetB2-based static baseline trained on
NAPS-processed sequences reaches a Macro F1 of 80.98\%, comparable to the
textual baseline of 80.55\%.

At the visual modeling level, DS-TANet integrates an EfficientNetB2 static
appearance stream with a frozen RAFT dynamic motion stream through motion-guided attention and Bi-GRU temporal modeling. Results confirm that static appearance and inter-frame motion are complementary, and that temporal modeling is essential for translating motion-guided modulation into effective sentiment representations. DS-TANet achieves a Macro F1 of 82.58\%.

At the multimodal fusion level, comparison of concatenation fusion and
Cross-Modal Attention shows that greater model complexity does not reliably
yield performance gains under limited training data. Without the acoustic
modality or additional data, DS-TAFNet achieves 87.49\% Acc2 and 87.48\%
Macro F1, outperforming all trimodal baselines in this study.

The central finding is that unimodal representation quality should be
prioritized before introducing complex cross-modal interaction---a practical
design principle for multimodal pattern recognition under data-limited
conditions. Beyond these contributions, NAPS can be reused as a modular
visual front end; DS-TANet's dual-stream design offers a reference for
strengthening weak visual modalities; and the data-scale-aware fusion
comparison provides guidance for lightweight fusion selection when labeled
data are limited.

Three limitations remain, each pointing to a direction for future work. First, although neither NAPS nor DS-TANet relies on language-specific information---their effectiveness stemming mainly from visual-level designs such as face localization, motion modeling, and temporal encoding---applicability to other language settings still requires validation with adapted text encoders; the transferability of both modules will therefore be evaluated on English and multilingual datasets, together with cross-dataset pretraining and data augmentation, to reassess the potential of high-capacity fusion structures such as Cross-Modal Attention under larger data scales. Second, the prosodic and intonational information contained in the acoustic modality may still complement visual expressive cues, so the acoustic modality will be incorporated once its representation quality is sufficiently improved, forming a complete trimodal system to investigate its complementarity with the optimized visual representations. Third, a small number of complex samples still require manual review, leaving room for improvement in automatic localization; NAPS will therefore be further optimized through methods such as audio-visual speaker verification to improve automatic localization accuracy, reduce manual review costs, and enhance the practical deployment feasibility of the proposed method.
\section*{Generative AI Disclosure}
During the preparation of this manuscript, the authors used generative AI tools for language refinement and for generating simple icons used in the figures. All research content, figures, and analyses are the authors' own. The authors reviewed and edited all AI-assisted output and take full responsibility for the published article.
\section*{CRediT authorship contribution statement}
Su Li: Writing – original draft, Visualization, Validation, Methodology, Data curation, Conceptualization, Investigation.
Yigong Zhang: Conceptualization, Writing – original draft, Writing – Review \& Editing, Supervision, Formal analysis, Funding acquisition.
Lei Xiong: Writing – Review \& Editing, Visualization, Formal analysis.
Chune Li: Writing – Review \& Editing, Visualization, Formal analysis.
\section*{Declaration of competing interest}
The authors declare that they have no known competing financial interests or personal relationships that could have appeared to influence the work reported in this paper.
\section*{Acknowledgments}
This research work is financially supported by the National Natural Science Foundation of China (grant Nos. 12173085) and the Training Object Project of technological innovation talents in Yunnan Province(No. 202305AD160004). 
\section*{Data availability}
Data will be made available on request.


\begin{thebibliography}{00}

\bibitem{das2023survey}
R. Das, T.D. Singh,
Multimodal sentiment analysis: A survey of methods, trends, and challenges,
\textit{ACM Computing Surveys} 55~(13s) (2023) Article 311,
doi:10.1145/3586075.

\bibitem{wang2023tetfn}
D. Wang, X. Guo, Y. Tian, J. Liu, L. He, X. Luo,
TETFN: A text enhanced transformer fusion network
for multimodal sentiment analysis,
\textit{Pattern Recognition}
136 (2023) 109259,
doi:10.1016/j.patcog.2022.109259.

\bibitem{peng2022ogm}
X. Peng, Y. Wei, A. Deng, D. Wang, D. Hu,
Balanced multimodal learning via on-the-fly gradient modulation,
in: \textit{Proceedings of the IEEE/CVF Conference on Computer Vision
and Pattern Recognition},
2022, pp. 8238--8247,
doi:10.1109/CVPR52688.2022.00806.

\bibitem{liu2022chsimsv2}
Y. Liu, Z. Yuan, H. Mao, Z. Liang, W. Yang, Y. Qiu,
T. Cheng, X. Li, H. Xu, K. Gao,
Make acoustic and visual cues matter: CH-SIMS v2.0 dataset
and AV-Mixup consistent module,
in: \textit{Proceedings of the 2022 International Conference
on Multimodal Interaction},
2022, pp. 268--279,
doi:10.1145/3536221.3556630.



\bibitem{zadeh2018mosei}
A.B. Zadeh, P.P. Liang, S. Poria, E. Cambria, L.-P. Morency,
Multimodal language analysis in the wild: CMU-MOSEI dataset
and interpretable dynamic fusion graph,
in: \textit{Proceedings of the 56th Annual Meeting of the
Association for Computational Linguistics, Volume 1: Long Papers},
2018, pp. 2236--2246.

\bibitem{yu2020ch}
W. Yu, H. Xu, F. Meng, Y. Zhu, Y. Ma, J. Wu, J. Zou, K. Yang,
CH-SIMS: A Chinese multimodal sentiment analysis dataset
with fine-grained annotation of modality,
in: \textit{Proceedings of the 58th Annual Meeting of the
Association for Computational Linguistics},
2020, pp. 3718--3727.





\bibitem{tan2019efficientnet}
M. Tan, Q.V. Le,
EfficientNet: Rethinking model scaling for convolutional neural networks,
in: \textit{Proceedings of the 36th International Conference
on Machine Learning},
2019, pp. 6105--6114.

\bibitem{liu2023expression}
Y. Liu, W. Wang, C. Feng, H. Zhang, Z. Chen, Y. Zhan,Expression snippet transformer for robust video-based facial expression recognition,
\textit{Pattern Recognition}
138 (2023) 109368,
doi:10.1016/j.patcog.2023.109368.

\bibitem{li2023multiscale}
T. Li, K.-L. Chan, T. Tjahjadi,
Multi-scale correlation module for video-based facial expression
recognition in the wild,
\textit{Pattern Recognition} 142 (2023) 109691,
doi:10.1016/j.patcog.2023.109691.

\bibitem{gan2025context}
Y. Gan, L. Xu, S. Song, X. Tao,
Context transformer with multiscale fusion for robust
facial emotion recognition,
\textit{Pattern Recognition}
167 (2025) 111720,
doi:10.1016/j.patcog.2025.111720.

\bibitem{teed2020raft}
Z. Teed, J. Deng,
RAFT: Recurrent all-pairs field transforms for optical flow,
in: \textit{Computer Vision -- ECCV 2020},
2020, pp. 402--419,
doi:10.1007/978-3-030-58536-5\_24.

\bibitem{simonyan2014twostream}
K. Simonyan, A. Zisserman,
Two-stream convolutional networks for action recognition in videos,
\textit{Advances in Neural Information Processing Systems}
27 (2014).

\bibitem{liu2026temporal}
F. Liu, B. Nan, X. Qian, X. Fu,Temporal--spatial cross-fusion for dynamic micro-expression recognition,
\textit{Pattern Recognition}
179 (2026) 113715,
doi:10.1016/j.patcog.2026.113715.

\bibitem{tsai2019multimodal}
Y.-H.H. Tsai, S. Bai, P.P. Liang, J.Z. Kolter,
L.-P. Morency, R. Salakhutdinov,
Multimodal transformer for unaligned multimodal language sequences,
in: \textit{Proceedings of the 57th Annual Meeting of the
Association for Computational Linguistics},
2019, pp. 6558--6569.

\bibitem{wei2026visual}
Q. Wei, Y. Zhou, J. Zhou, L. Ye, Y. Zhang,
Visual label augmentation-driven multimodal emotion recognition,
\textit{Pattern Recognition}
179 (2026) 113886,
doi:10.1016/j.patcog.2026.113886.

\bibitem{devlin2019bert}
J. Devlin, M.-W. Chang, K. Lee, K. Toutanova,
BERT: Pre-training of deep bidirectional transformers
for language understanding,
in: \textit{Proceedings of the 2019 Conference of the North American
Chapter of the Association for Computational Linguistics:
Human Language Technologies, Volume 1: Long and Short Papers},
2019, pp. 4171--4186.

\bibitem{cui2020macbert}
Y. Cui, W. Che, T. Liu, B. Qin, S. Wang, G. Hu,
Revisiting pre-trained models for Chinese natural language processing,
in: \textit{Findings of the Association for Computational Linguistics:
EMNLP 2020},
2020, pp. 657--668.

\bibitem{lugaresi2019mediapipe}
C. Lugaresi, J. Tang, H. Nash, C. McClanahan, E. Uboweja,
M. Hays, F. Zhang, C.-L. Chang, M.G. Yong, J. Lee,
W.-T. Chang, W. Hua, M. Georg, M. Grundmann,
MediaPipe: A framework for perceiving and processing reality,
in: \textit{Third Workshop on Computer Vision for AR/VR at CVPR},
2019.

\bibitem{cao2018vggface2}
Q. Cao, L. Shen, W. Xie, O.M. Parkhi, A. Zisserman,
VGGFace2: A dataset for recognising faces across pose and age,
in: \textit{2018 13th IEEE International Conference on Automatic
Face \& Gesture Recognition},
2018,
doi:10.1109/FG.2018.00020.



\bibitem{vanma2024mot}
L. Van Ma, T.T.D. Nguyen, C. Shim, D.Y. Kim, N. Ha, M. Jeon,
Visual multi-object tracking with re-identification and occlusion
handling using labeled random finite sets,
\textit{Pattern Recognition} 156 (2024) 110785,
doi:10.1016/j.patcog.2024.110785.

\bibitem{zou2025unsupervised}
Z. Zou, D. Jia, W. Tang,
Towards unsupervised learning of joint facial landmark detection
and head pose estimation,
\textit{Pattern Recognition}
162 (2025) 111393,
doi:10.1016/j.patcog.2025.111393.

\bibitem{wan2024precise}
J. Wan, H. Liu, Y. Wu, Z. Lai, W. Min, J. Liu,
Precise facial landmark detection by Dynamic Semantic Aggregation Transformer,
\textit{Pattern Recognition}
156 (2024) 110827,
doi:10.1016/j.patcog.2024.110827.

\bibitem{mollahosseini2019affectnet}
A. Mollahosseini, B. Hasani, M.H. Mahoor,
AffectNet: A database for facial expression, valence, and arousal
computing in the wild,
\textit{IEEE Transactions on Affective Computing}
10~(1) (2019) 18--31,
doi:10.1109/TAFFC.2017.2740923.

\bibitem{cho2014gru}
K. Cho, B. van Merri{\"e}nboer, C. Gulcehre, D. Bahdanau,
F. Bougares, H. Schwenk, Y. Bengio,
Learning phrase representations using RNN encoder--decoder
for statistical machine translation,
in: \textit{Proceedings of the 2014 Conference on Empirical
Methods in Natural Language Processing},
2014, pp. 1724--1734,
doi:10.3115/v1/D14-1179.

\bibitem{liu2023pretrain}
P. Liu, W. Yuan, J. Fu, Z. Jiang, H. Hayashi, G. Neubig,
Pre-train, prompt, and predict: A systematic survey of prompting
methods in natural language processing,
\textit{ACM Computing Surveys}
55~(9) (2023) 195:1--195:35,
doi:10.1145/3560815.

\bibitem{vaswani2017attention}
A. Vaswani, N. Shazeer, N. Parmar, J. Uszkoreit,
L. Jones, A.N. Gomez, {\L}. Kaiser, I. Polosukhin,
Attention is all you need,
\textit{Advances in Neural Information Processing Systems}
30 (2017) 5998--6008.



\bibitem{lin2017focal}
T.-Y. Lin, P. Goyal, R. Girshick, K. He, P. Doll{\'a}r,
Focal loss for dense object detection,
in: \textit{Proceedings of the IEEE International Conference
on Computer Vision},
2017, pp. 2980--2988,
doi:10.1109/ICCV.2017.324.

\bibitem{goodfellow2015explaining}
I.J. Goodfellow, J. Shlens, C. Szegedy,
Explaining and harnessing adversarial examples,
in: \textit{International Conference on Learning Representations},
2015.

\bibitem{williams2018}
J. Williams, S. Kleinegesse, R. Comanescu, O. Radu,
Recognizing emotions in video using multimodal DNN feature fusion,
in: \textit{Proceedings of Grand Challenge and Workshop on Human
Multimodal Language},
2018, pp. 11--19.

\bibitem{zadeh2017}
A. Zadeh, M. Chen, S. Poria, E. Cambria, L.-P. Morency,
Tensor fusion network for multimodal sentiment analysis,
in: \textit{Proceedings of the 2017 Conference on Empirical
Methods in Natural Language Processing},
2017, pp. 1103--1114,
doi:10.18653/v1/D17-1115.

\bibitem{liu2018}
Z. Liu, Y. Shen, V.B. Lakshminarasimhan, P.P. Liang,
A. Zadeh, L.-P. Morency,
Efficient low-rank multimodal fusion with modality-specific factors,
in: \textit{Proceedings of the 56th Annual Meeting of the
Association for Computational Linguistics},
2018, pp. 2247--2256,
doi:10.18653/v1/P18-1209.

\bibitem{zadeh2018mfn}
A. Zadeh, P.P. Liang, N. Mazumder, S. Poria,
E. Cambria, L.-P. Morency,
Memory fusion network for multi-view sequential learning,
in: \textit{Proceedings of the AAAI Conference on Artificial Intelligence}
32~(1) (2018) 5634--5641,
doi:10.1609/aaai.v32i1.12021.

\bibitem{rahman2020}
W. Rahman, M.K. Hasan, S. Lee, A.B. Zadeh,
C. Mao, L.-P. Morency, E. Hoque,
Integrating multimodal information in large pretrained transformers,
in: \textit{Proceedings of the 58th Annual Meeting of the
Association for Computational Linguistics},
2020, pp. 2359--2369,
doi:10.18653/v1/2020.acl-main.214.

\bibitem{hazarika2020}
D. Hazarika, R. Zimmermann, S. Poria,
MISA: Modality-invariant and modality-specific representations
for multimodal sentiment analysis,
in: \textit{Proceedings of the 28th ACM International Conference
on Multimedia},
2020, pp. 1122--1131,
doi:10.1145/3394171.3413678.

\bibitem{han2021}
W. Han, H. Chen, S. Poria,
Improving multimodal fusion with hierarchical mutual information
maximization for multimodal sentiment analysis,
in: \textit{Proceedings of the 2021 Conference on Empirical
Methods in Natural Language Processing},
2021, pp. 9180--9192,
doi:10.18653/v1/2021.emnlp-main.723.

\bibitem{yu2021}
W. Yu, H. Xu, Z. Yuan, J. Wu,
Learning modality-specific representations with self-supervised
multi-task learning for multimodal sentiment analysis,
in: \textit{Proceedings of the AAAI Conference on Artificial Intelligence}
35~(12) (2021) 10790--10797,
doi:10.1609/aaai.v35i12.17289.

\bibitem{zhong2025cmc}
G. Zhong, J. Li, H. Zhu, R. Huan, Y. Pan,
Calibrating multimodal consensus for emotion recognition,
\textit{arXiv preprint arXiv:2510.20256} (2025).

\bibitem{sun2024emt}
L. Sun, Z. Lian, B. Liu, J. Tao,
Efficient multimodal transformer with dual-level feature restoration
for robust multimodal sentiment analysis,
\textit{IEEE Transactions on Affective Computing}
15~(1) (2024) 309--325,
doi:10.1109/TAFFC.2022.3226530.

\bibitem{luo2025triaged}
Y. Luo, W. Liu, Q. Sun, S. Li, J. Li, R. Wu, X. Tang,
TriagedMSA: Triaging sentimental disagreement in multimodal
sentiment analysis,
\textit{IEEE Transactions on Affective Computing} (2025),
doi:10.1109/TAFFC.2024.3524789.

\bibitem{yuan2024mcteacher}
Z. Yuan, J. Fang, H. Xu, K. Gao,
Multimodal consistency-based teacher for semi-supervised
multimodal sentiment analysis,
\textit{IEEE/ACM Transactions on Audio, Speech, and Language Processing}
32 (2024) 3413--3425,
doi:10.1109/TASLP.2024.3422792.

\bibitem{zhang2026multimodal}
Q. Zhang, Y. Sui, H. Guo, J. Liu, J. Duan, H. Wang,
L. He, L. Song, G. Xu,
Multimodal emotion recognition via large model guided dialogue
state tracking with dynamic graph refinement,
\textit{Pattern Recognition}
180 (2026) 114018,
doi:10.1016/j.patcog.2026.114018.
\end{thebibliography}



\end{document}